\documentclass[letterpaper, 10 pt, conference]{ieeeconf}  

\IEEEoverridecommandlockouts                              

\usepackage{times}
\usepackage{epsfig}
\usepackage{graphicx}
\usepackage{amsmath}
\usepackage{amssymb}

\usepackage{fancybox}
\usepackage{amsfonts,amsmath,amssymb,latexsym}
\usepackage{array}
\usepackage{graphicx}

\newenvironment{separate}{\begin{quote}\begin{sffamily}\footnotesize}{\end{sffamily}\end{quote}}

\newcommand{\beq}{\begin{equation}}
\newcommand{\eeq}{\end{equation}}
\newcommand{\beqa}{\begin{eqnarray}}
\newcommand{\eeqa}{\end{eqnarray}}
\newcommand{\beqas}{\begin{eqnarray*}}
\newcommand{\eeqas}{\end{eqnarray*}}

\newlength{\boxwidth}
\newlength{\narrowboxwidth}
\usepackage{upquote}

\usepackage[pagebackref=true,breaklinks=true,letterpaper=true,colorlinks,bookmarks=false]{hyperref}

\newcommand{\drive}{\texttt{DR(eye)VE }}

\title{\LARGE \bf
Learning Where to Attend Like a Human Driver
}

\author{Andrea Palazzi*, Francesco Solera*, Simone Calderara, Stefano Alletto, Rita Cucchiara 
\thanks{* Equal contribution}%
\thanks{Authors are with the Department of Computer Engineering, University of Modena and Reggio Emilia, Italy
        {\tt\small name.surname@unimore.it}}%
}

\begin{document}

\maketitle
\thispagestyle{empty}
\pagestyle{empty}

\begin{abstract}
Despite the advent of autonomous cars, it's likely - at least in the near future - that human attention will still maintain a central role as a guarantee in terms of legal responsibility during the driving task.
In this paper we study the dynamics of the driver's gaze and use it as a proxy to understand related attentional mechanisms.
First, we build our analysis upon two questions: \emph{where} and \emph{what} the driver is looking at?
Second, we model the driver's gaze by training a coarse-to-fine convolutional network on short sequences extracted from the \drive dataset.
Experimental comparison against different baselines reveal that the driver's gaze can indeed be learnt to some extent, despite i) being highly subjective and ii) having only one driver's gaze available for each sequence due to the irreproducibility of the scene.
Eventually, we advocate for a new assisted driving paradigm which suggests to the driver, with no intervention, where she should focus her attention.
\end{abstract}

\section{Introduction}

While autonomous driving is quickly reaching maturity, it's not clear how far in time society will overlook the legal responsibility of the human driver~\cite{federalguidalines2016}.
Conversely, Advanced Driver Assistance Systems (ADAS) are human-centric and already established both in literature and in the market. 

\begin{figure}[t]
    \centering
    \begin{tabular}{cc}
    \hspace{-0.2cm}\includegraphics[width=0.24\textwidth]{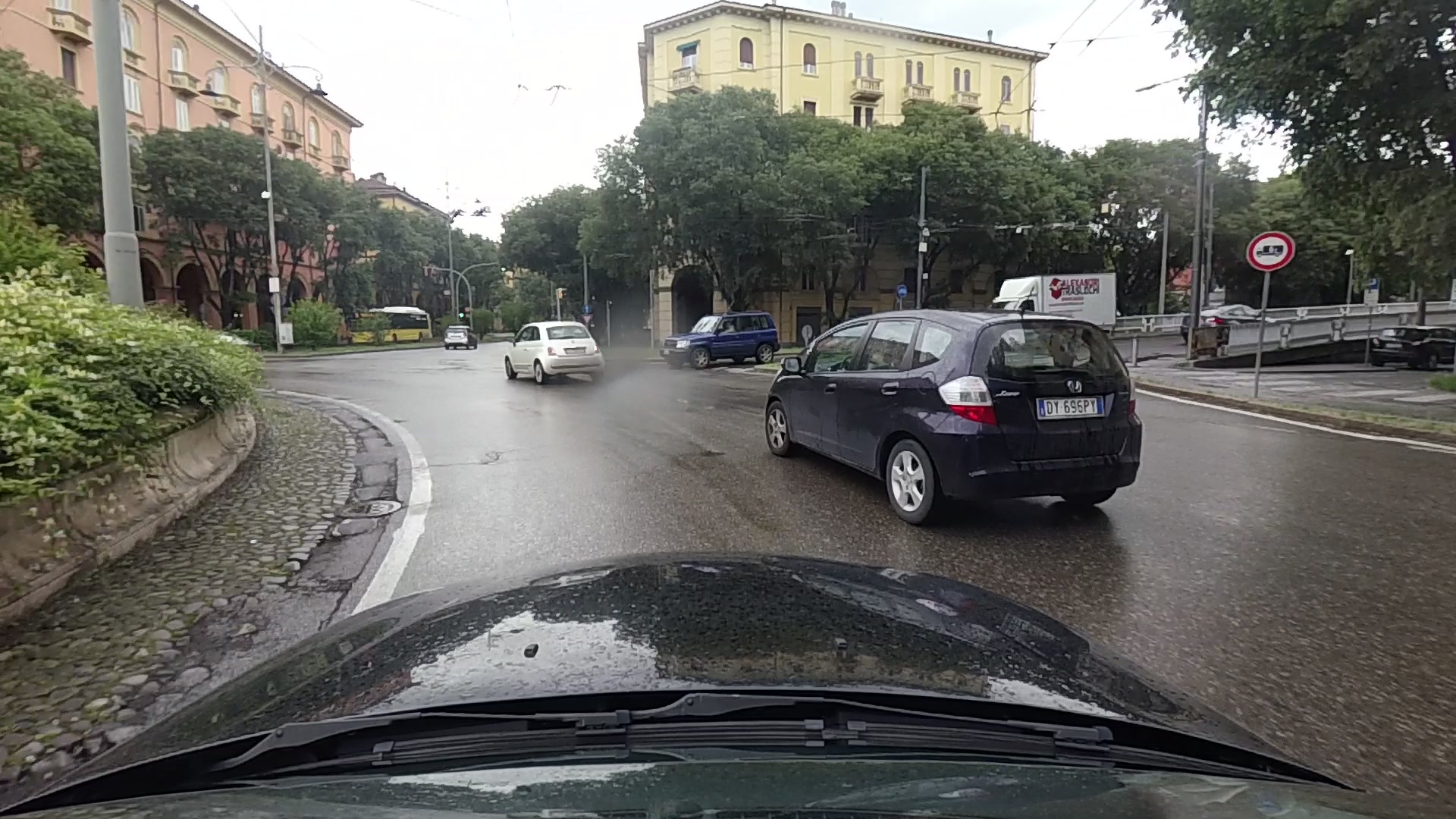} &\hspace{-0.3cm}\includegraphics[width=0.24\textwidth]{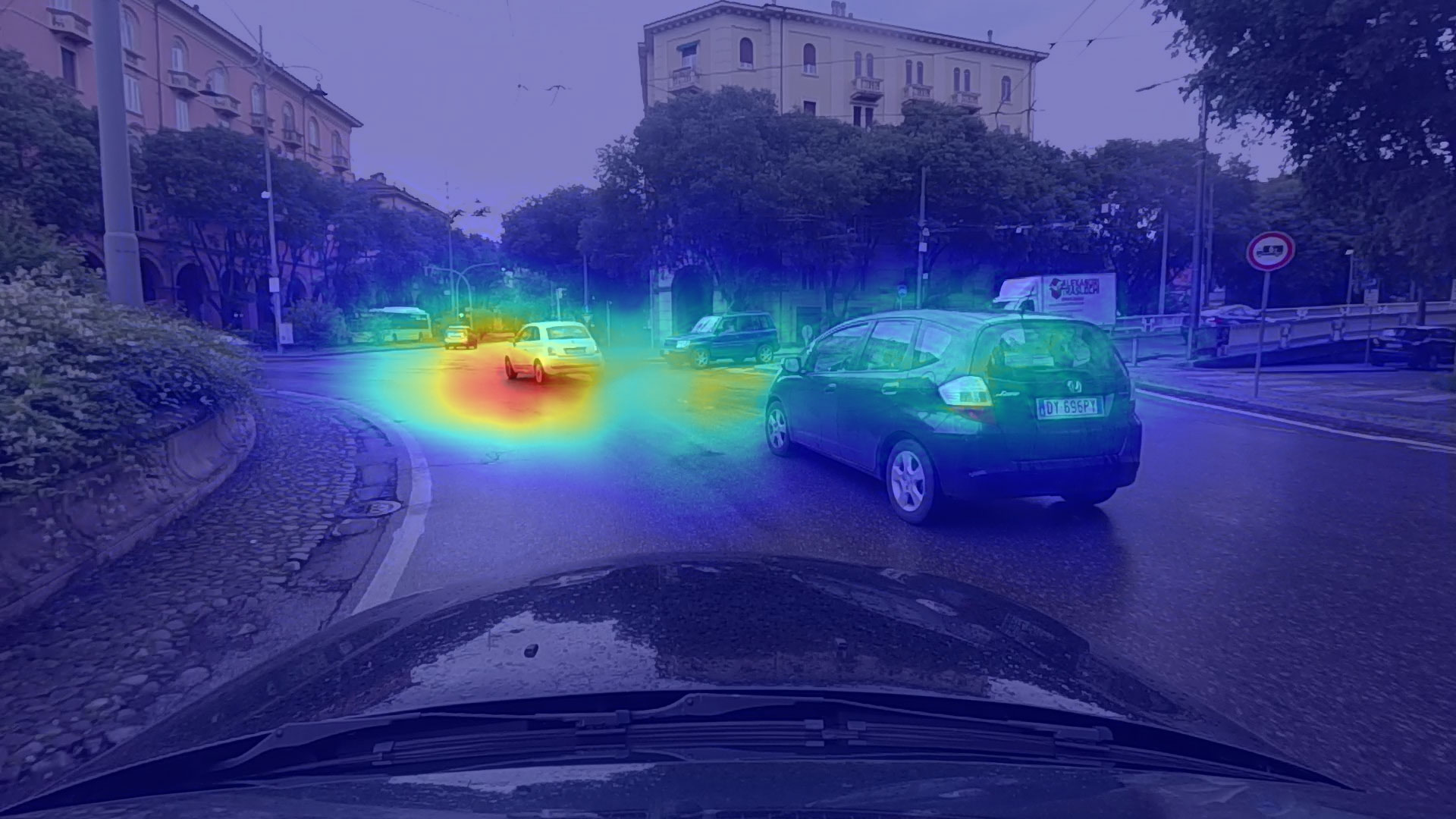}\\
    \hspace{-0.2cm}(a) roof-mounted camera & \hspace{-0.3cm}(b) network prediction \\
    \hspace{-0.2cm}\includegraphics[width=0.24\textwidth]{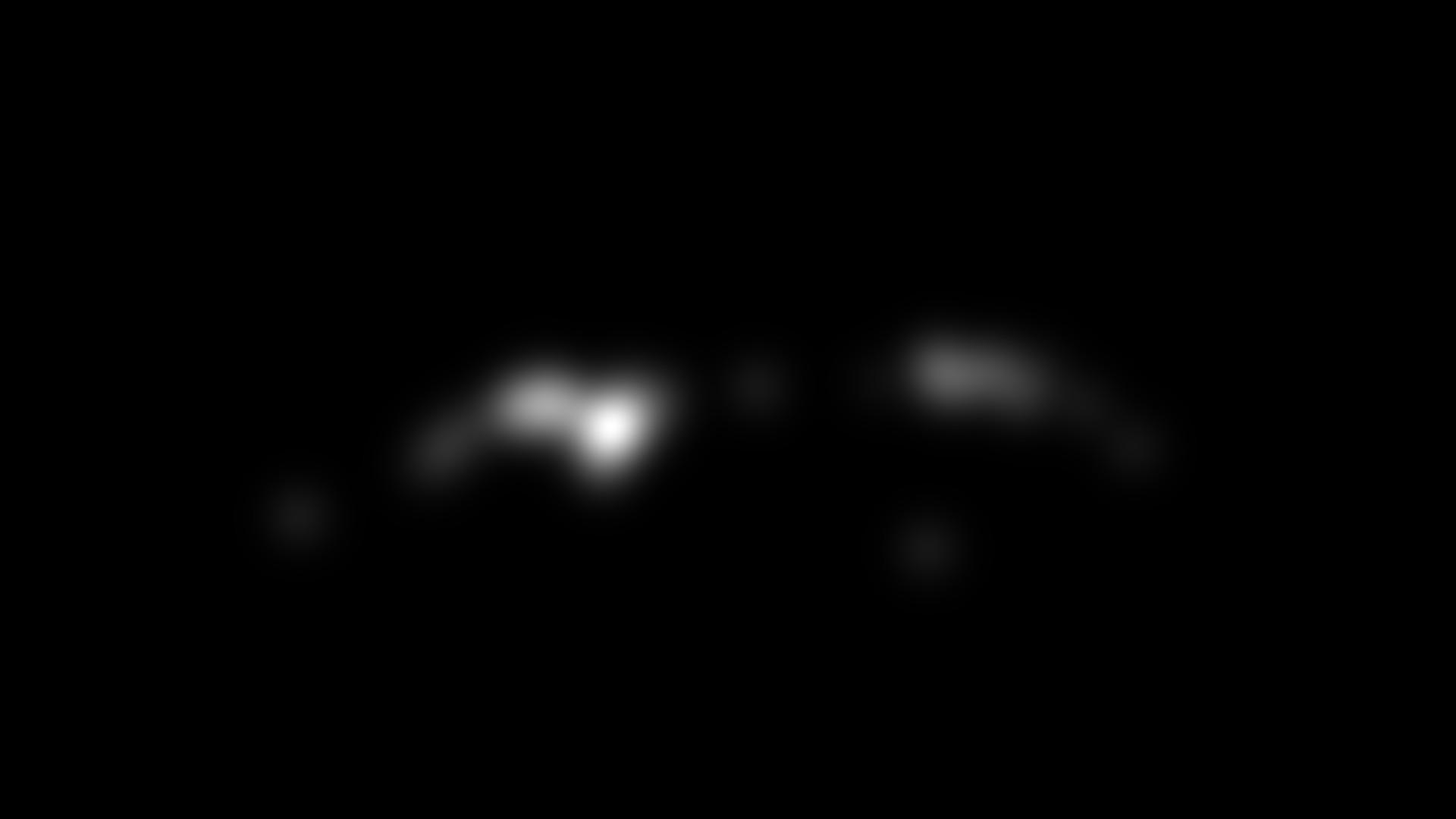} &\hspace{-0.3cm}\includegraphics[width=0.24\textwidth]{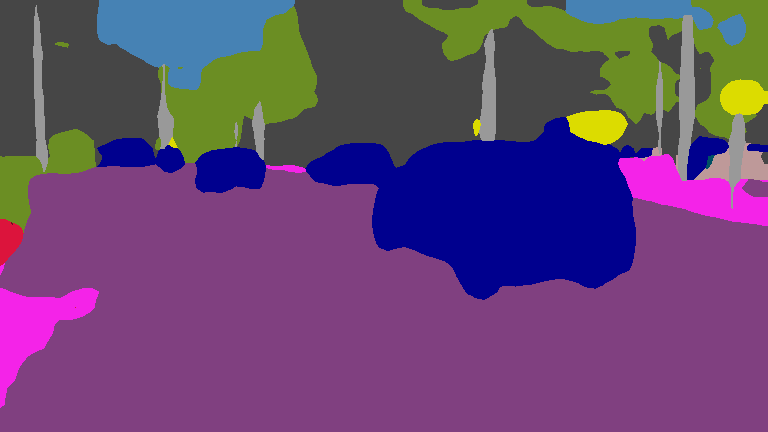}\\
    \hspace{-0.2cm}(c) attentional map (\emph{where}) & \hspace{-0.3cm}(d) segmentation (\emph{what})
    \end{tabular}
    \caption{Starting from raw frames and attentional maps provided with the \drive~\cite{alletto2016dr} dataset, we investigate \emph{where} the driver focuses (c) and \emph{what} does she look at (d). We then replicate the observed attentional behavior through a deep network model (b). }
    \label{fig:overview}
\end{figure}

Some of the most ambitious examples of assisted driving are related to driver monitoring systems~\cite{jain2015car,frolich2014will, kumar2013learning, morris2011lane}, where the attentional behavior of the driver is parsed together with the road scene to predict potentially unsafe manoeuvres and act on the car in order to avoid them (either by signaling the driver or braking).
However, all these approaches are limited by their ability to capture the true attentional and intentional behavior of the driver, which is still a complex and largely unsolved task today.
Conversely, we advocate for a new assisted driving paradigm which suggests to the driver, with no hard intervention, where she should focus her attention. The problem is thus shifted from a personal level (\emph{what the driver is looking at}) to a task-driven level (\emph{what the driver should be looking at}).
Following the notion that gaze is a primary cue to human visual attention, the contributions of this paper are twofold:
\begin{itemize}
    \item we investigate the attentional dynamics during the driving task by employing data collected from eye tracking glasses and semantic cues, Fig.~\ref{fig:overview}(c) and (d) respectively (Sec.~\ref{sec:dreyeve_dataset_analysis});
    \item upon these findings, we build a deep network architecture to model human attention while driving, Fig.~\ref{fig:overview}(b), and evaluate the ability of the proposed approach to replicate what we observed on humans (Sec.~\ref{sec:models}-\ref{sec:exp2}).
\end{itemize}
\noindent We train our model on the \drive data~\cite{alletto2016dr}, under the hypothesis that i) individuals were thoughtfully driving and ii) distractions have no predefined pattern and can thus be considered outliers. Given that the provided gaze annotation is reliable, we aim to generalize from \emph{what the driver was looking at} in annotated sequences to \emph{what the driver should be looking at} in unseen scenarios.

\noindent Code and pre-trained model are available at: \\
\begingroup
    \fontsize{9pt}{5pt}\selectfont
        \url{https://github.com/francescosolera/dreyeving}.
\endgroup

\begin{figure*}[t!]
    \centering
    \begin{tabular}{ccccc}
    \includegraphics[width=0.18\textwidth]{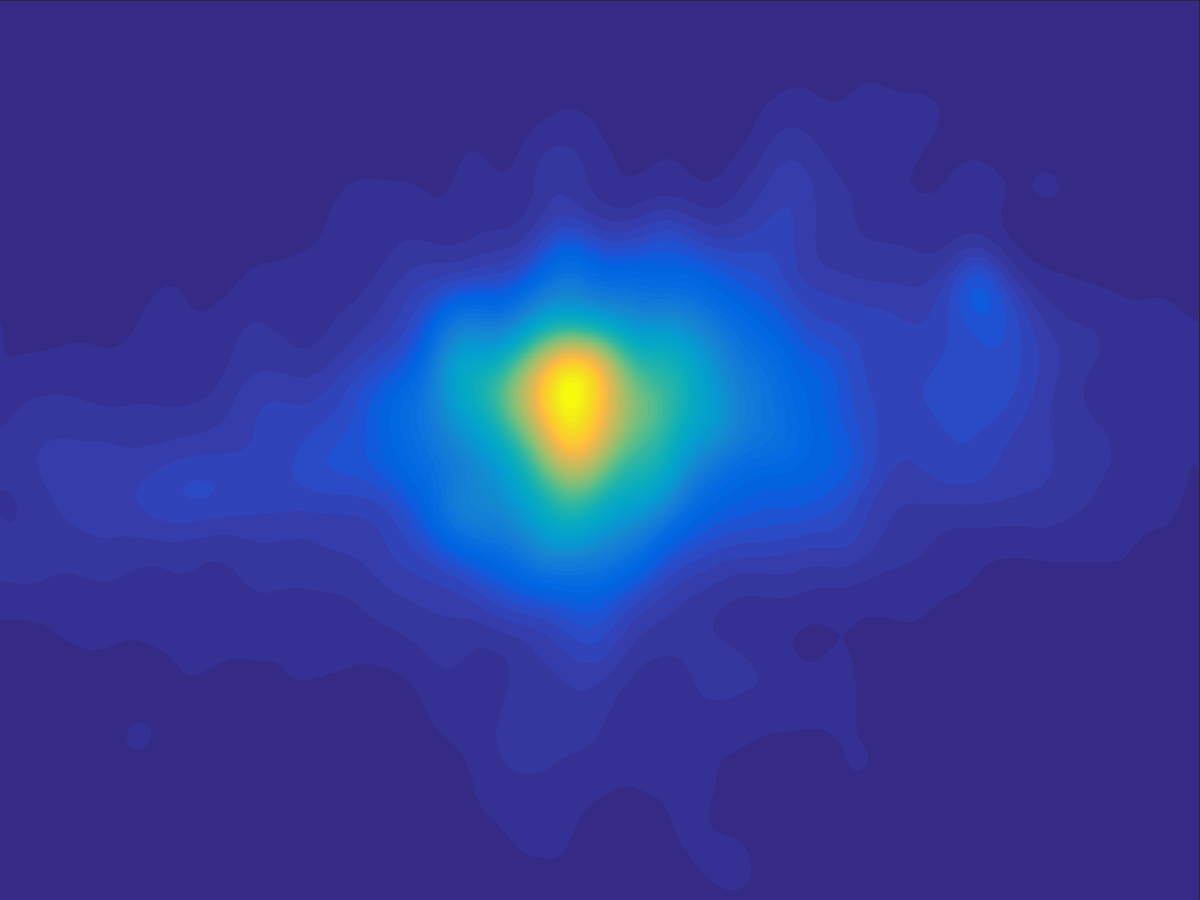} &
    \hspace{-0.2cm} \includegraphics[width=0.18\textwidth]{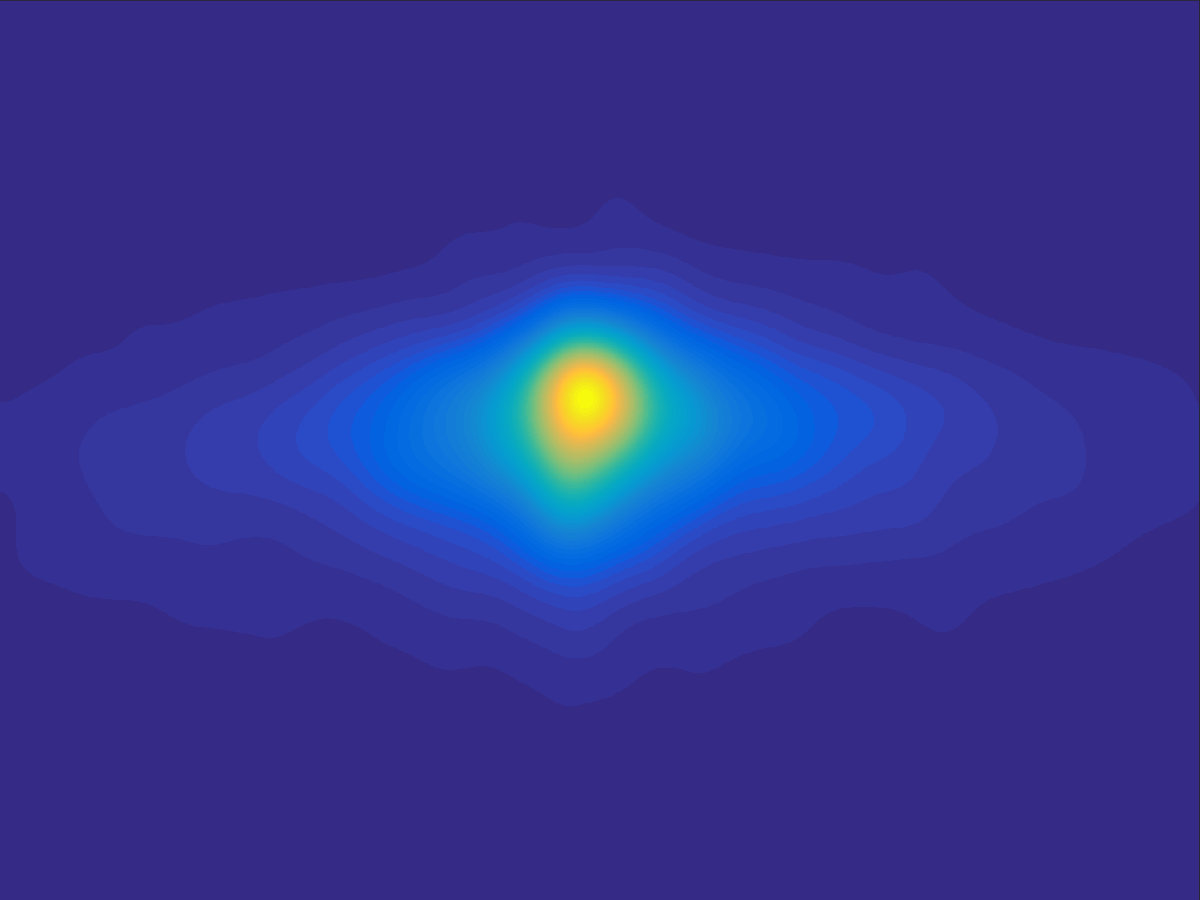} &
    \hspace{-0.2cm} \includegraphics[width=0.18\textwidth]{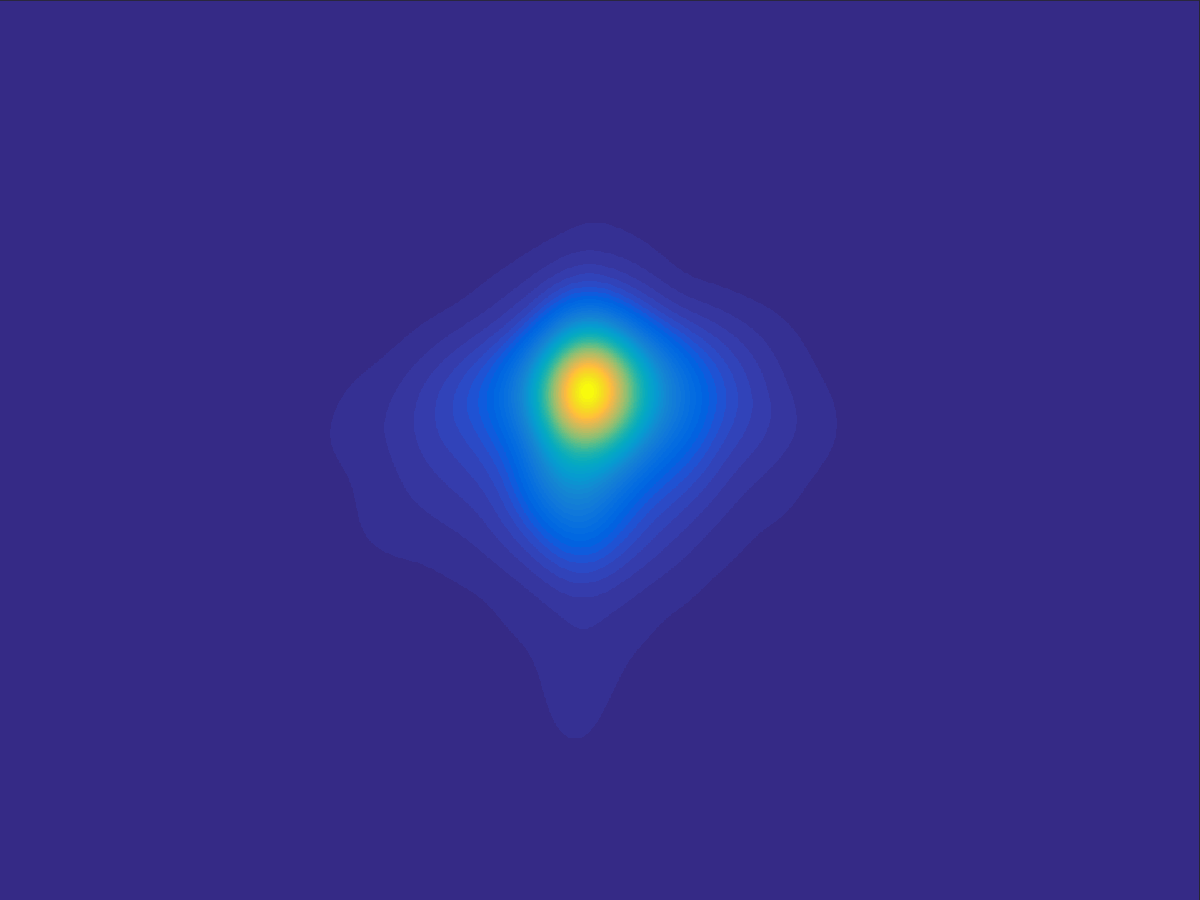} &
    \hspace{-0.2cm} \includegraphics[width=0.18\textwidth]{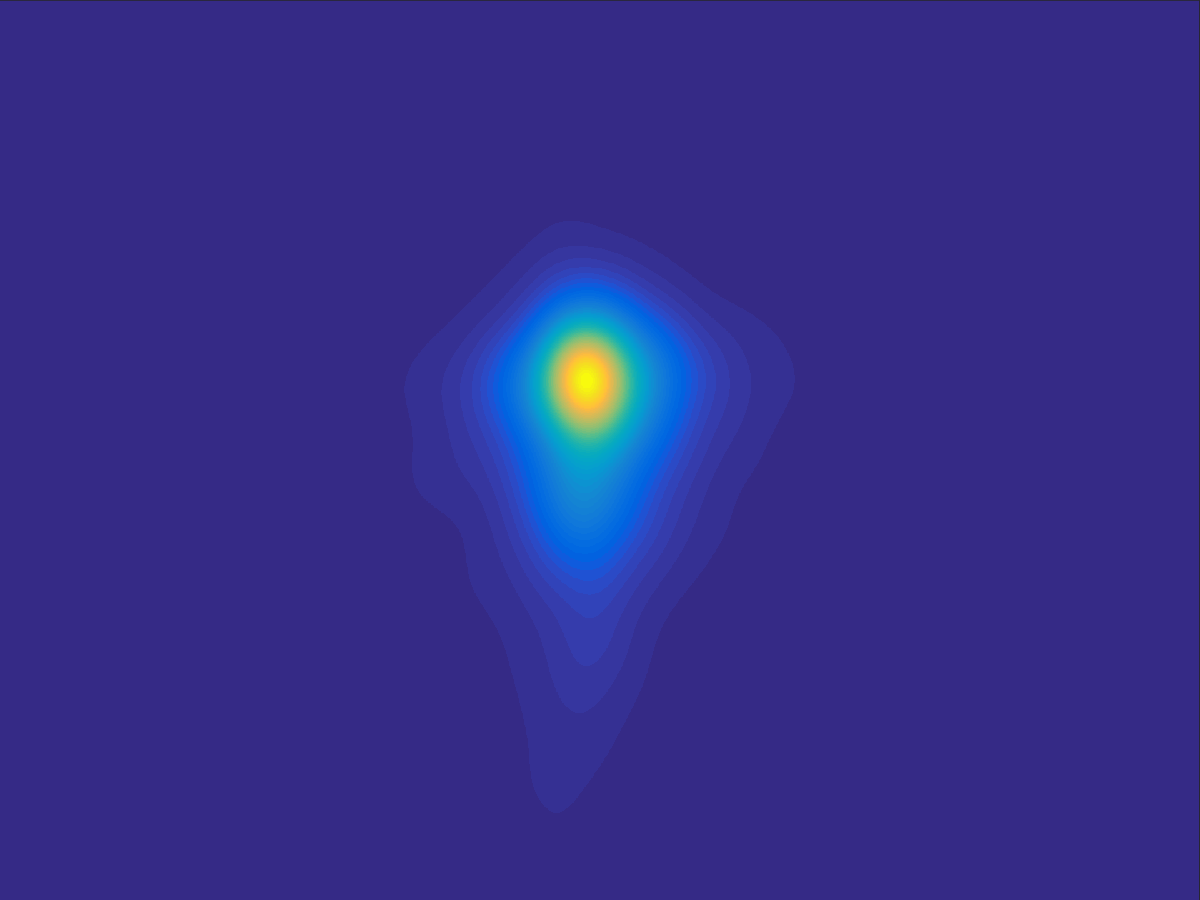} &
    \hspace{-0.2cm} \includegraphics[width=0.18\textwidth]{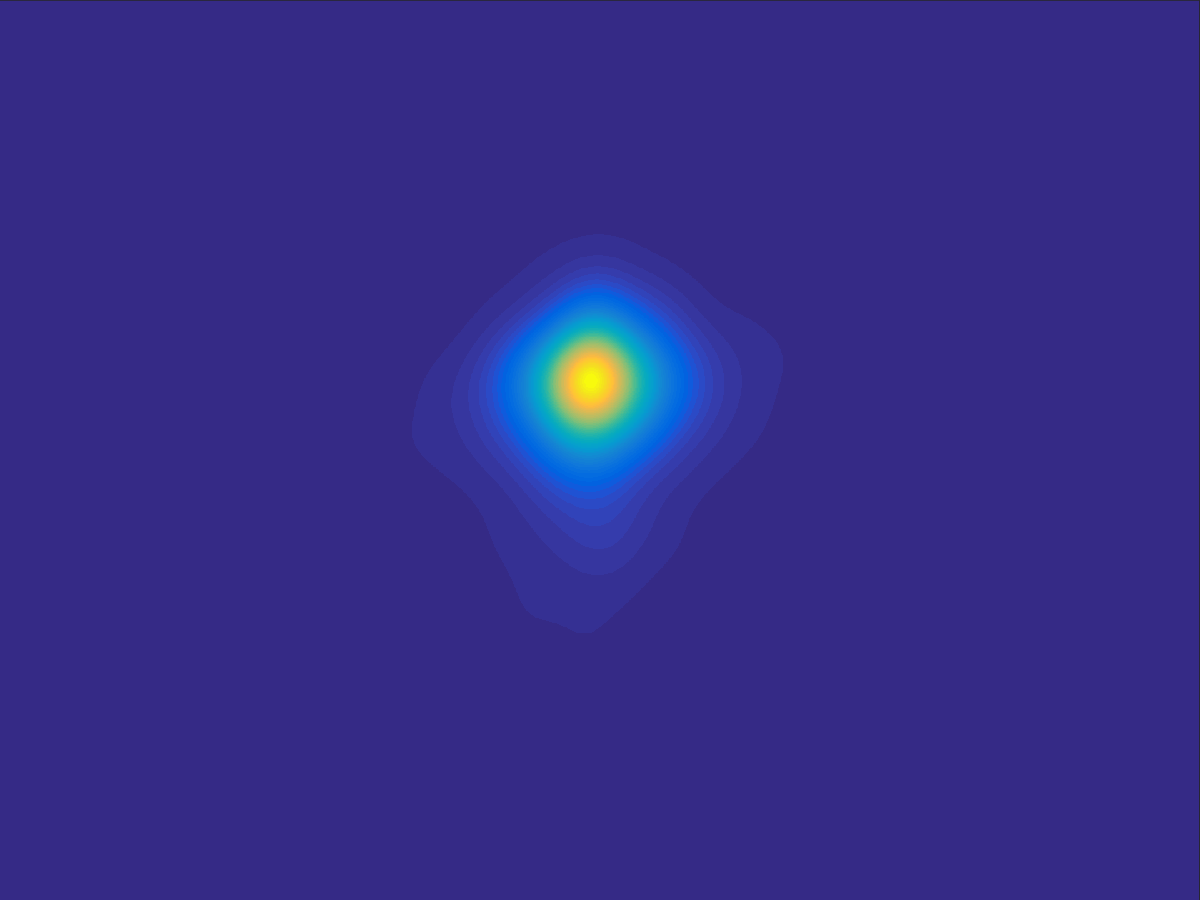}\\
    \includegraphics[width=0.18\textwidth, height=0.25cm]{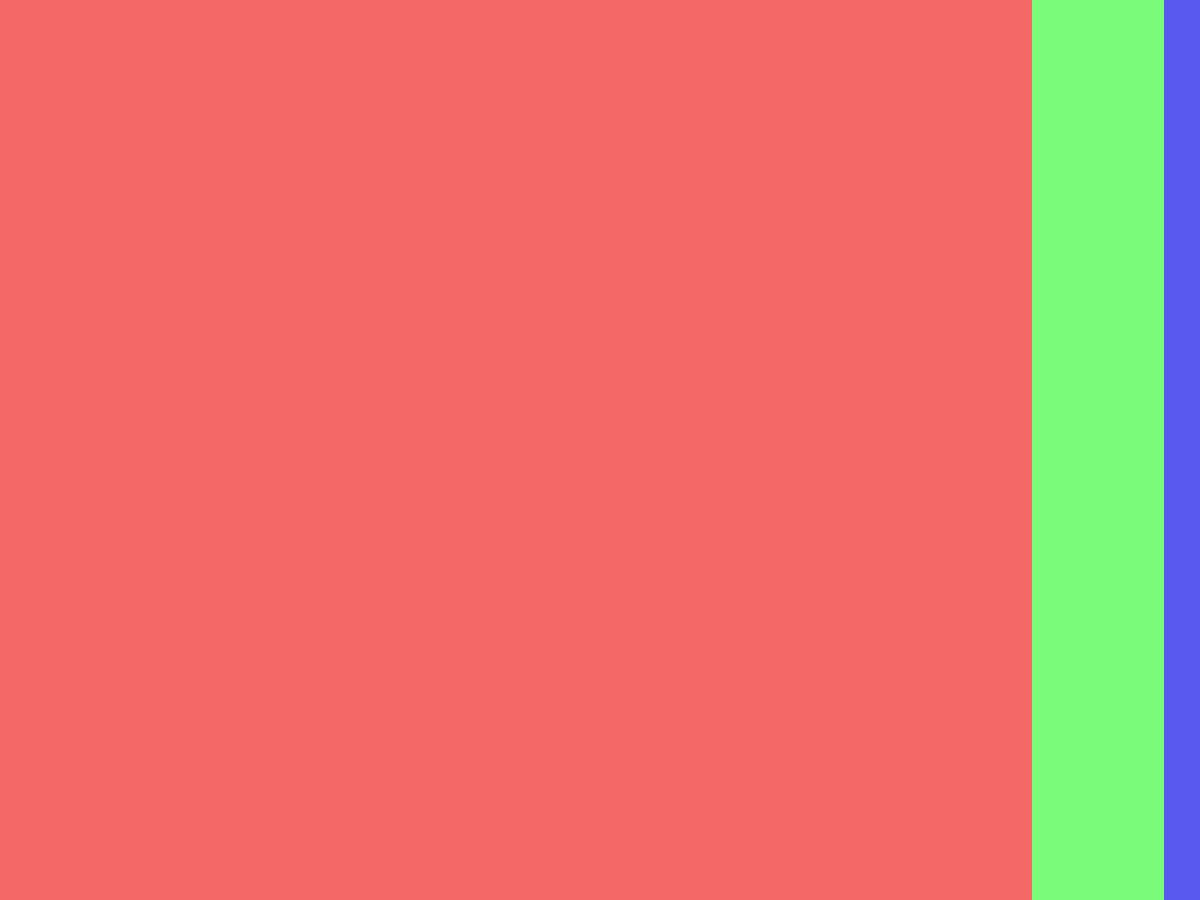} &
    \hspace{-0.2cm} \includegraphics[width=0.18\textwidth, height=0.25cm]{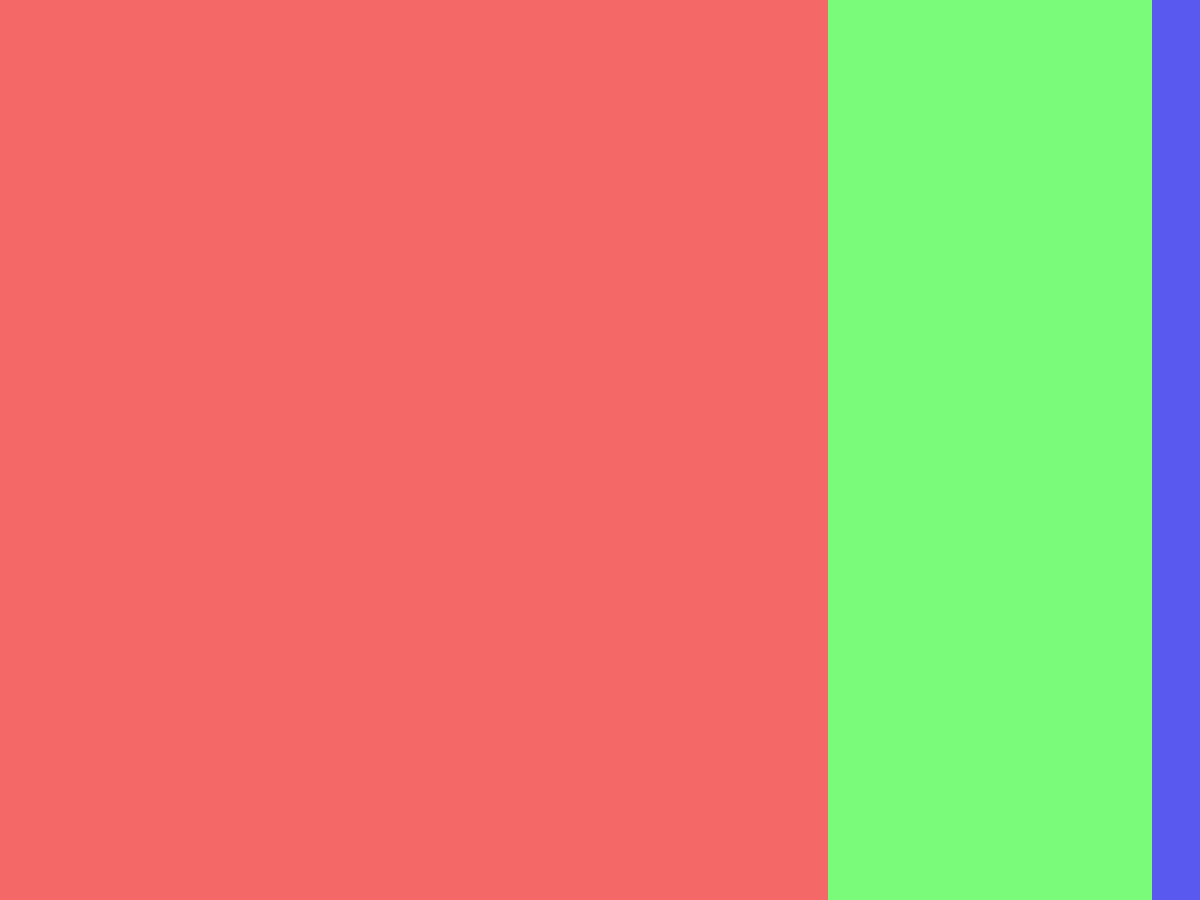} &
    \hspace{-0.2cm} \includegraphics[width=0.18\textwidth, height=0.25cm]{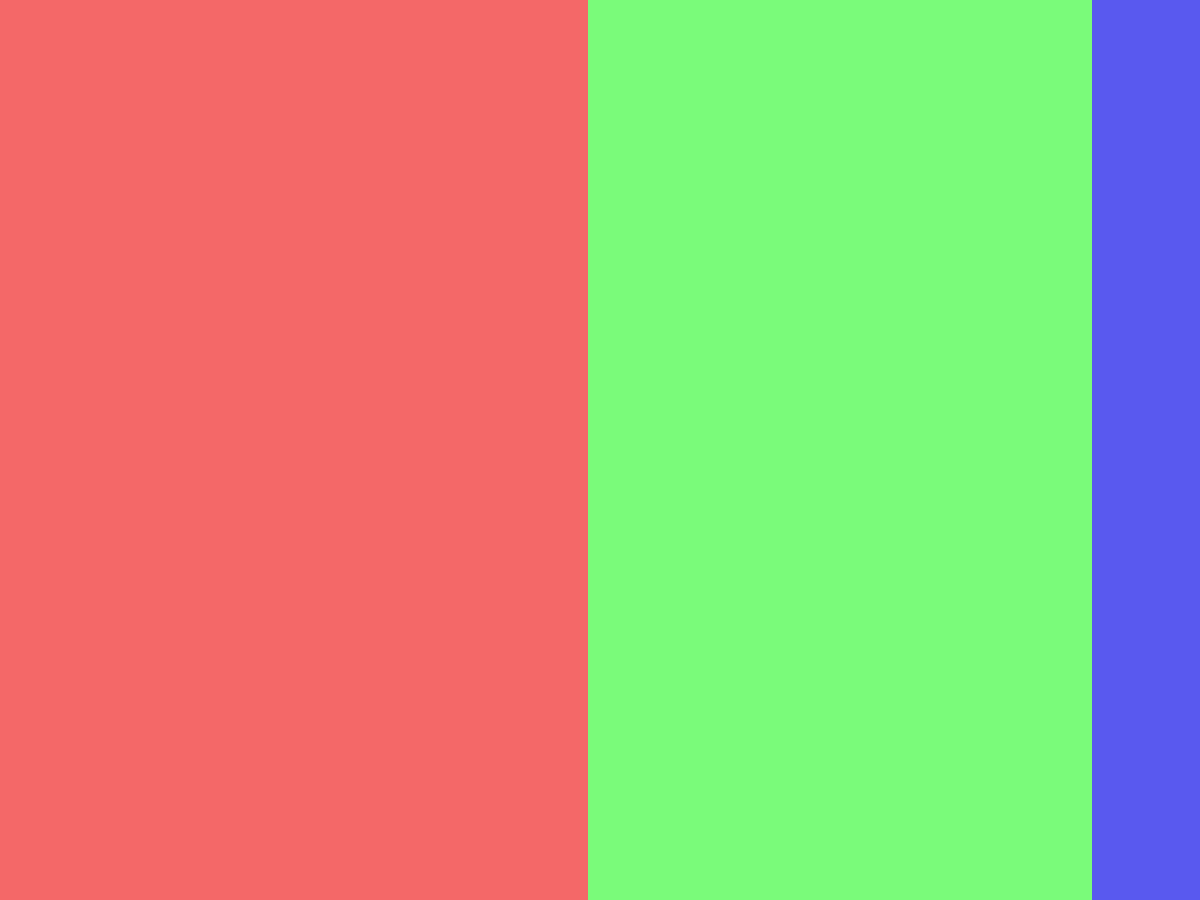} &
    \hspace{-0.2cm} \includegraphics[width=0.18\textwidth, height=0.25cm]{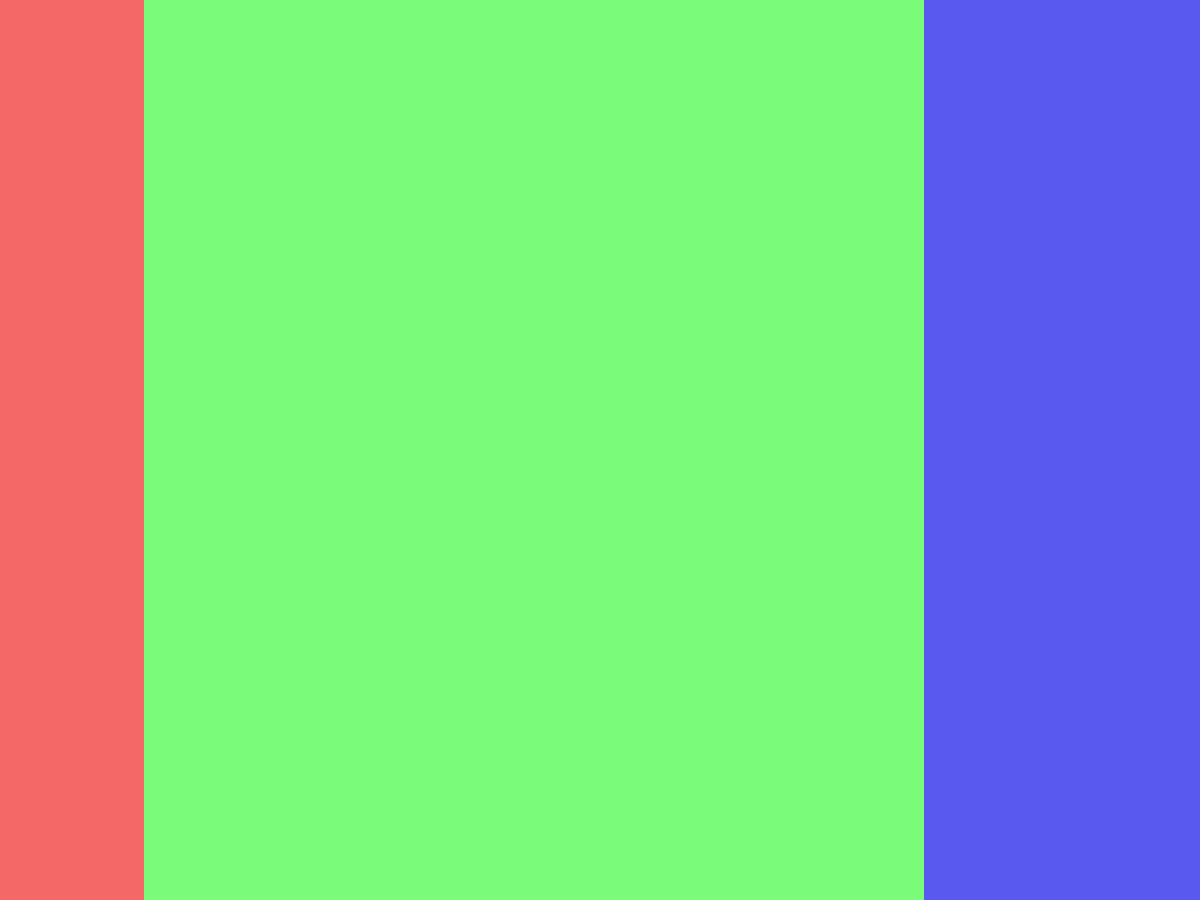} &
    \hspace{-0.2cm} \includegraphics[width=0.18\textwidth, height=0.25cm]{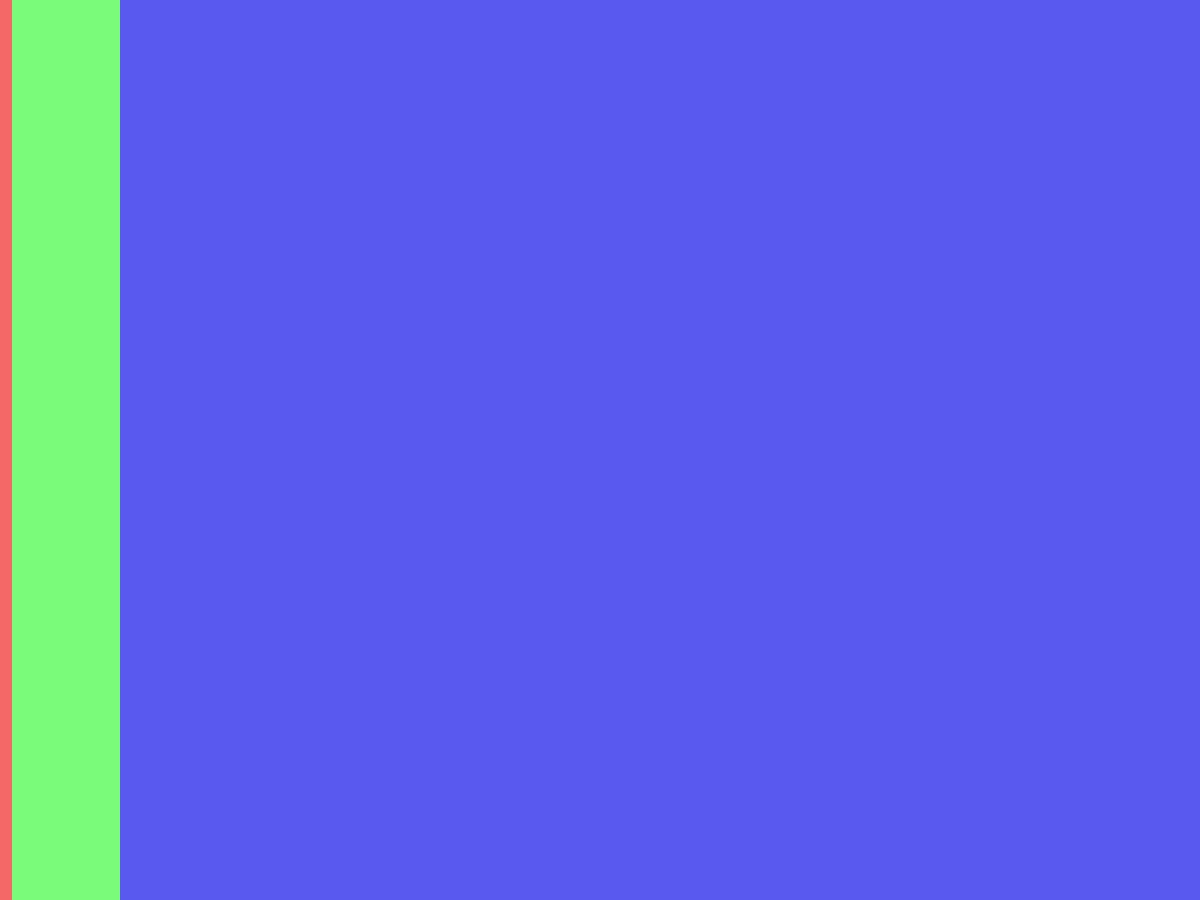}\\
    (a) $0\leq \mathrm{km/h}\leq10$ &
    (b) $10\leq \mathrm{km/h}\leq30$ &
    (c) $30\leq \mathrm{km/h}\leq50$ &
    (d) $50\leq \mathrm{km/h}\leq70$ &
    (e) $70\leq \mathrm{km/h}$
    \end{tabular}
    \caption{The driver's attention converges towards the vanishing point as the speed gradually increases. (a) When the car is almost stationary, the driver is distracted by various objects in the scene. (b) At slightly higher speeds, the focus extends towards the sides of the frame: indeed this is the typical speed at which we take on curves. (c-e) As the speed increases further, the driver's gaze deviates less and less from the vanishing point of the road. See Sec.~\ref{sec:where} for details about the red, green and blue bars. Best viewed on screen.}
    \label{fig:gt_across_speed}
\end{figure*}

\section{Related Works}
\noindent Our proposal relates to the following topics.\\
%
{\bf Head pose and gaze estimation.}
\cite{tawari2014driver, valenti2012combining, bar2012driver, mbouna2013visual}~estimate the driver's gaze direction using both head and eye location cues.
They aim to infer eyes off-the-road and other dangerous driving habits, as~\cite{tawari2014robust,vicente2015driver,fridman2015driver}.
Others exploit driver's attention to predict next maneuvers, and use either head-pose estimation and GPS/maps~\cite{jain2015car} or collect precise gaze data through eye tracking devices~\cite{lethaus2011using}.
Conversely, \cite{surrey808725} doesn't employ driver's attention information (neither measured nor estimated) to predict actions. They show a strict correlation between the visual appearance of the scene and driver's action such as steering, turning or braking. This finding also motivates our work, where we try to model the driver's attention from the external scene only.\\
{\bf Video saliency.} In this work we model the driver's focus through attentional maps, which resemble video saliency maps. Video saliency is typically tackled by extracting low level (or bottom-up) features from each frame an unsupervised~\cite{mauthner2015encoding, wang2015saliency, wang2015consistent} or supervised~\cite{zhong2013video,6942210} manner. Temporal dependencies are enforced by conditioning the prediction on previous frames~\cite{rudoy2013learning} or by means of optical flow~\cite{zhong2013video,6942210}. To the best of our knowledge, this work is the first one to exploit deep learning to tackle video saliency.
%


\section{Experimental analysis of driver's attention}
\label{sec:dreyeve_dataset_analysis}
In this section we investigate the attentional mechanisms involved in driving. To this end, we rely on the recently proposed \drive~\cite{alletto2016dr} dataset, which is a collection of 74 sequences 5 minutes long, featuring different landscapes, weather scenarios, lighting conditions and 8 drivers.
Every sequence is composed of two videos, capturing both the driver's and the car point of view. While the driver's gaze is synchronized with the former video, ground truth attentional maps are provided on the latter through a homographic projection followed by spatial smoothing and time integration. 
These latter post-processing steps attenuate subjectivity of the drivers' scan-path and reduce measurement inaccuracies in the ground truth.
More formally, we can define an attentional map as a 2D grid where each cell represents the probability that the corresponding pixel of the roof-mounted camera image is within the driver's focus of attention.

\begin{figure}[b]
    \centering
    \begin{tabular}{cc}
        \hspace{-0.2cm}\includegraphics[width=0.236\textwidth]{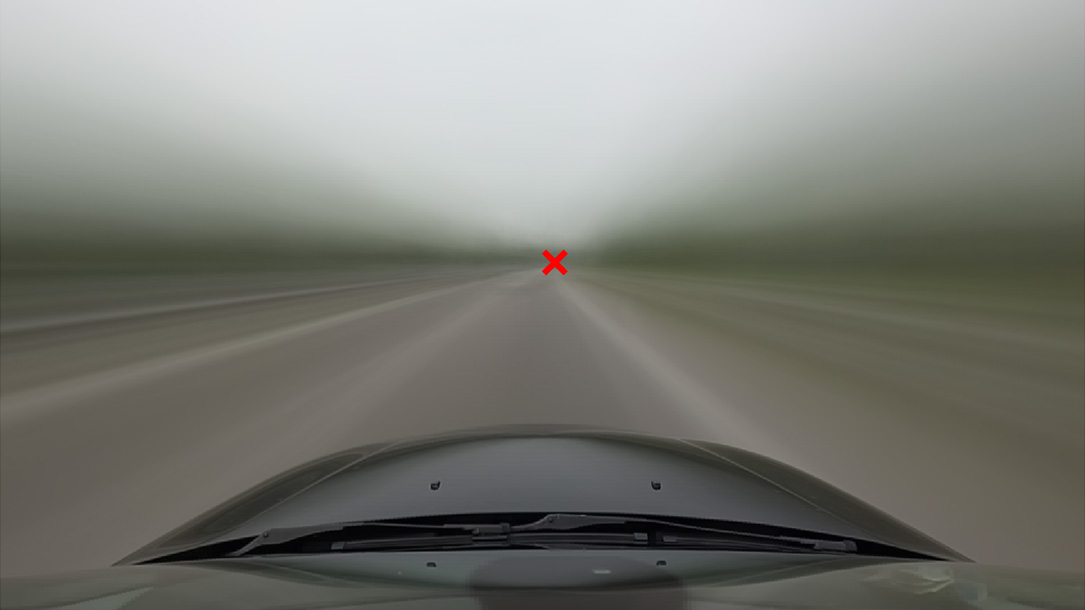} & \hspace{-0.3cm} \includegraphics[width=0.236\textwidth]{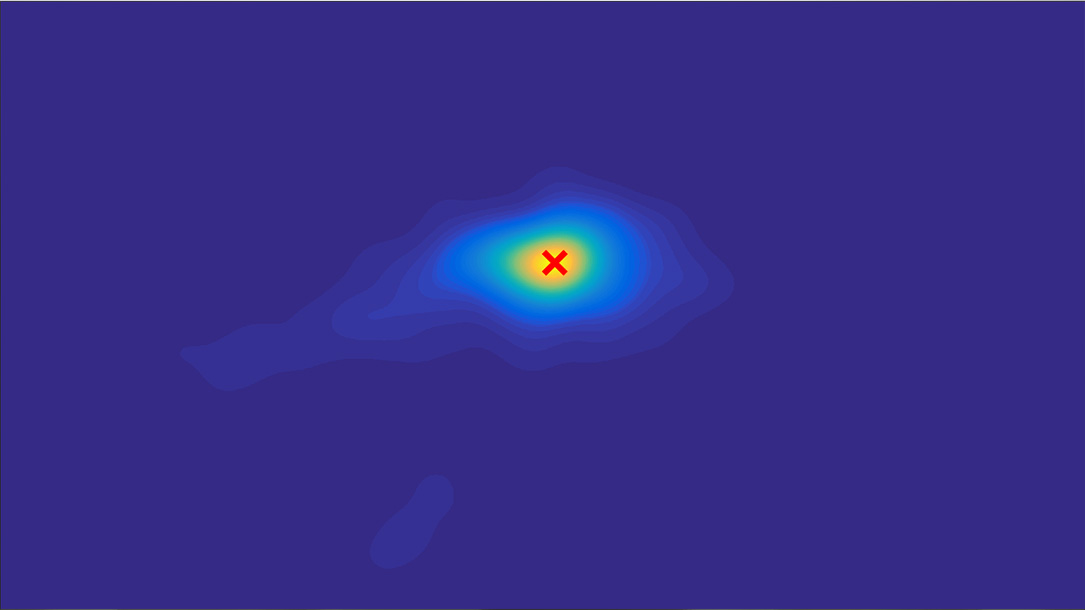} \\
        (a) & (b)
    \end{tabular}
    \caption{Frames (a) and attentional maps (b) averaged across the whole sequence \texttt{02}. The red cross marks the mode of all fixations for this sequence, qualitatively highlighting the link between driver's focus and the vanishing point of the road.}
    \label{fig:vanishing_point}
\end{figure}

\subsection{Where do we attend while driving?}
\label{sec:where}


\begin{figure*}[t!]
    \begin{tabular}{cccccccccc}
      \includegraphics[height=3cm]{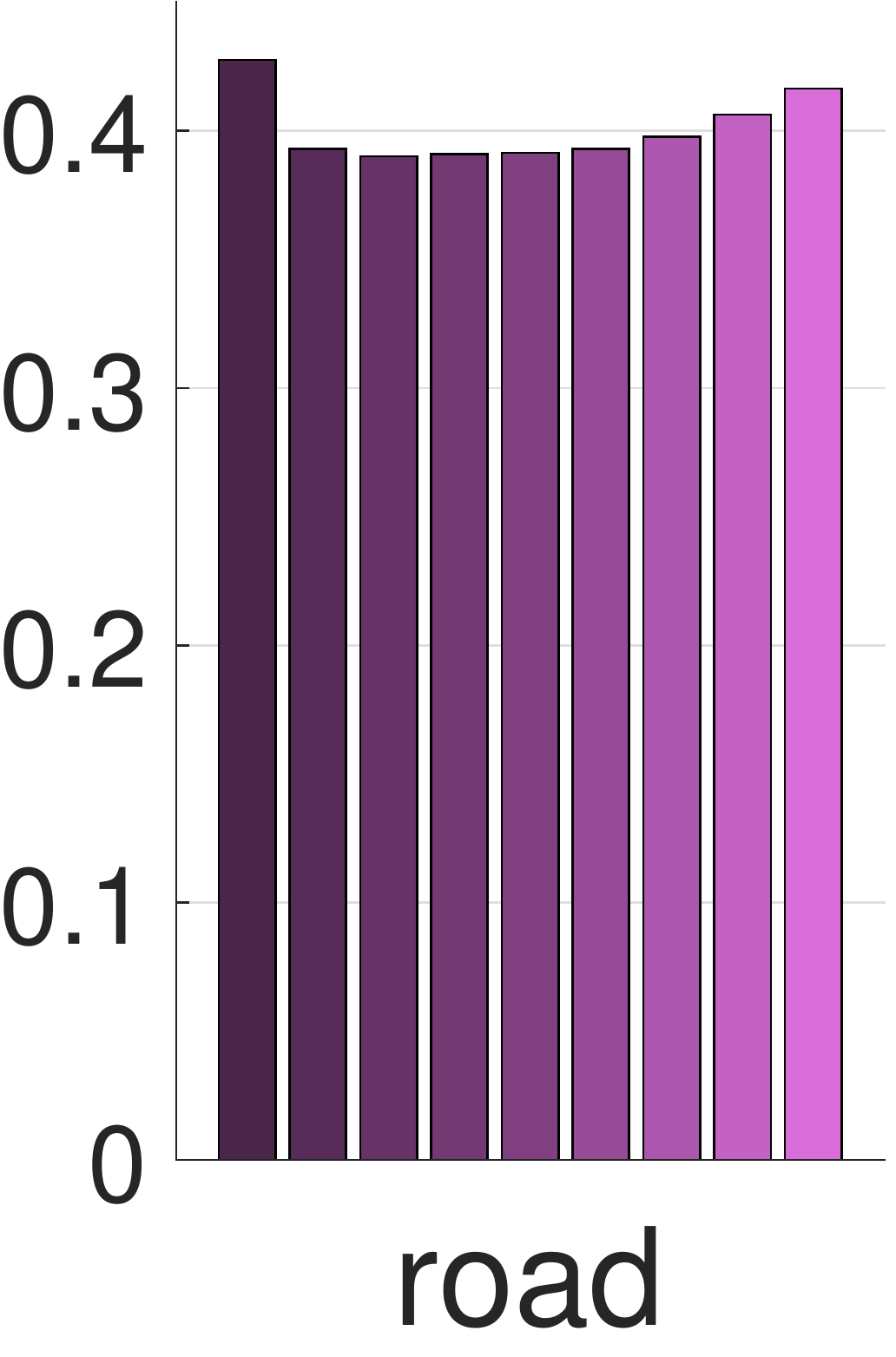} &
      \hspace{-0.5cm}\includegraphics[height=3cm]{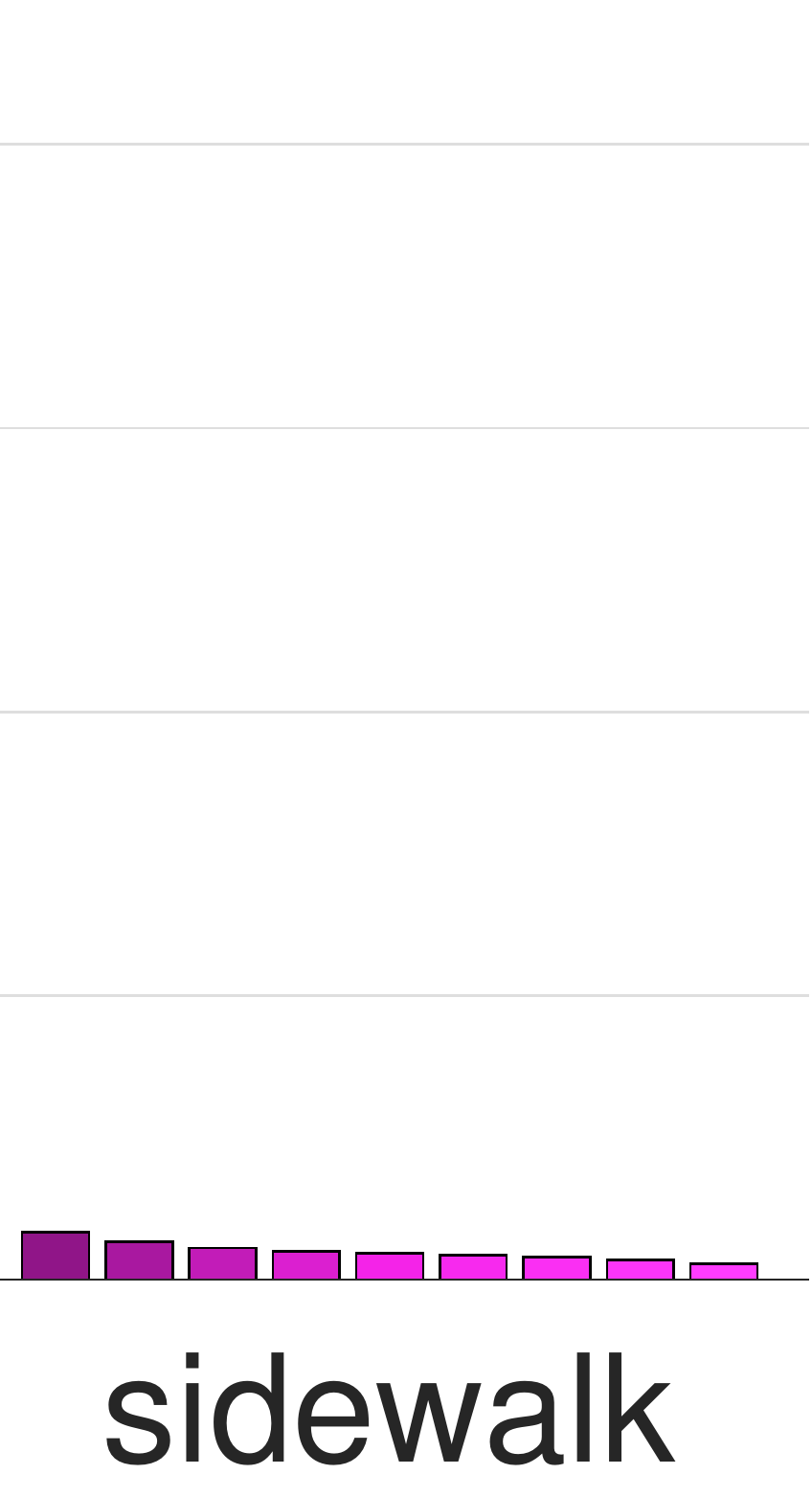} &
      \hspace{-0.45cm}\includegraphics[height=3cm]{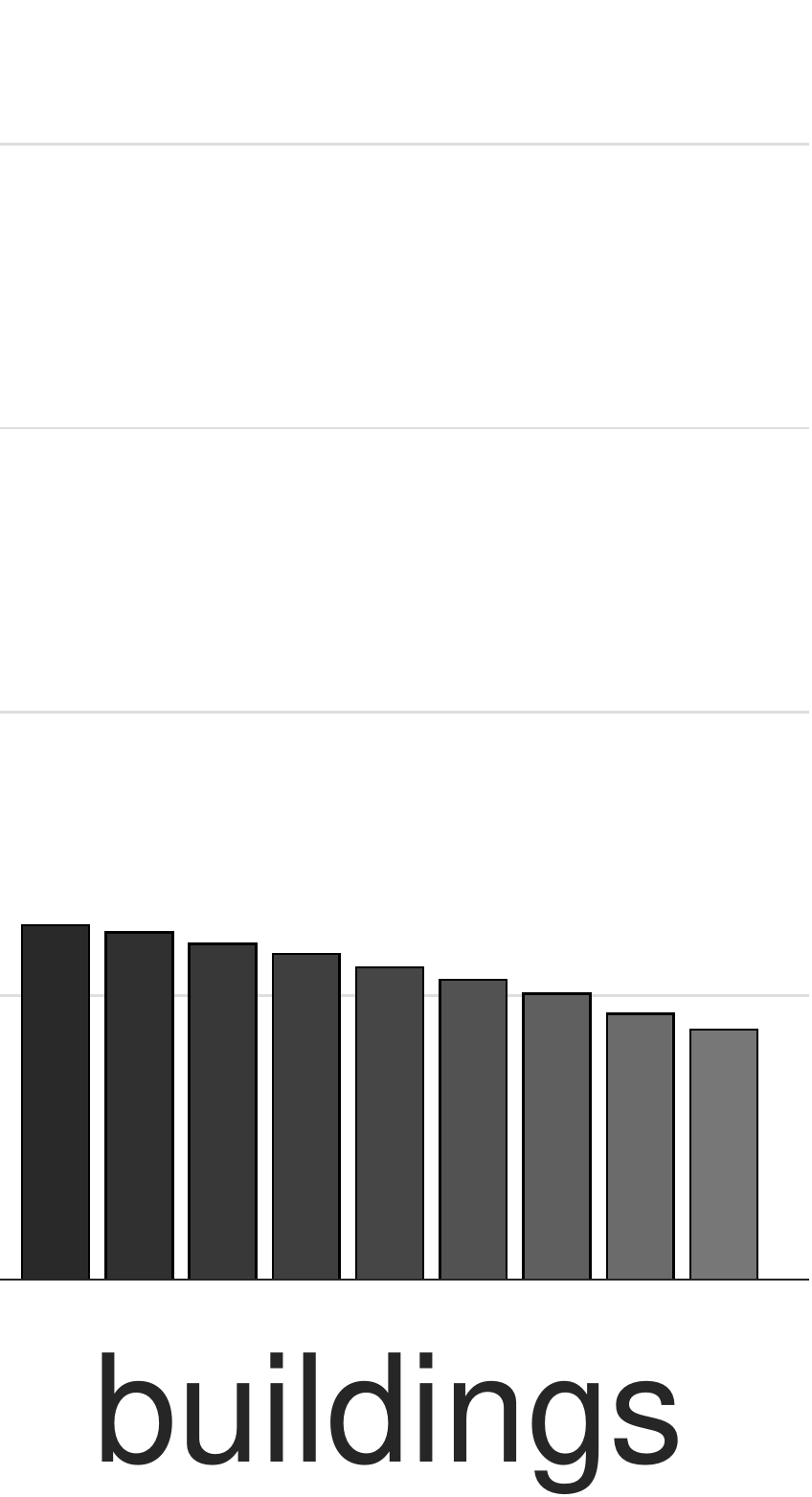} &
      \hspace{-0.45cm}\includegraphics[height=3cm]{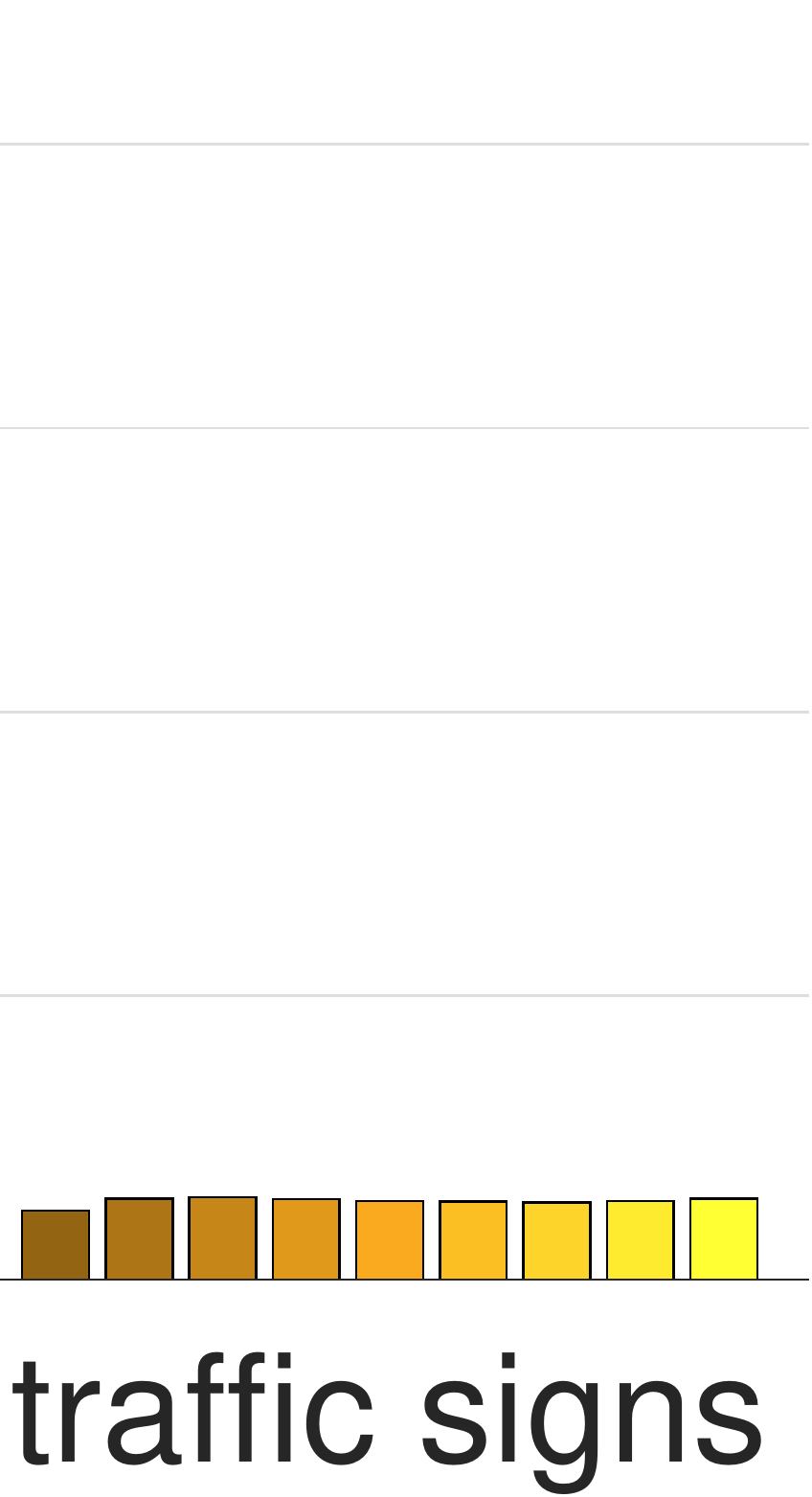} &
      \hspace{-0.45cm}\includegraphics[height=3cm]{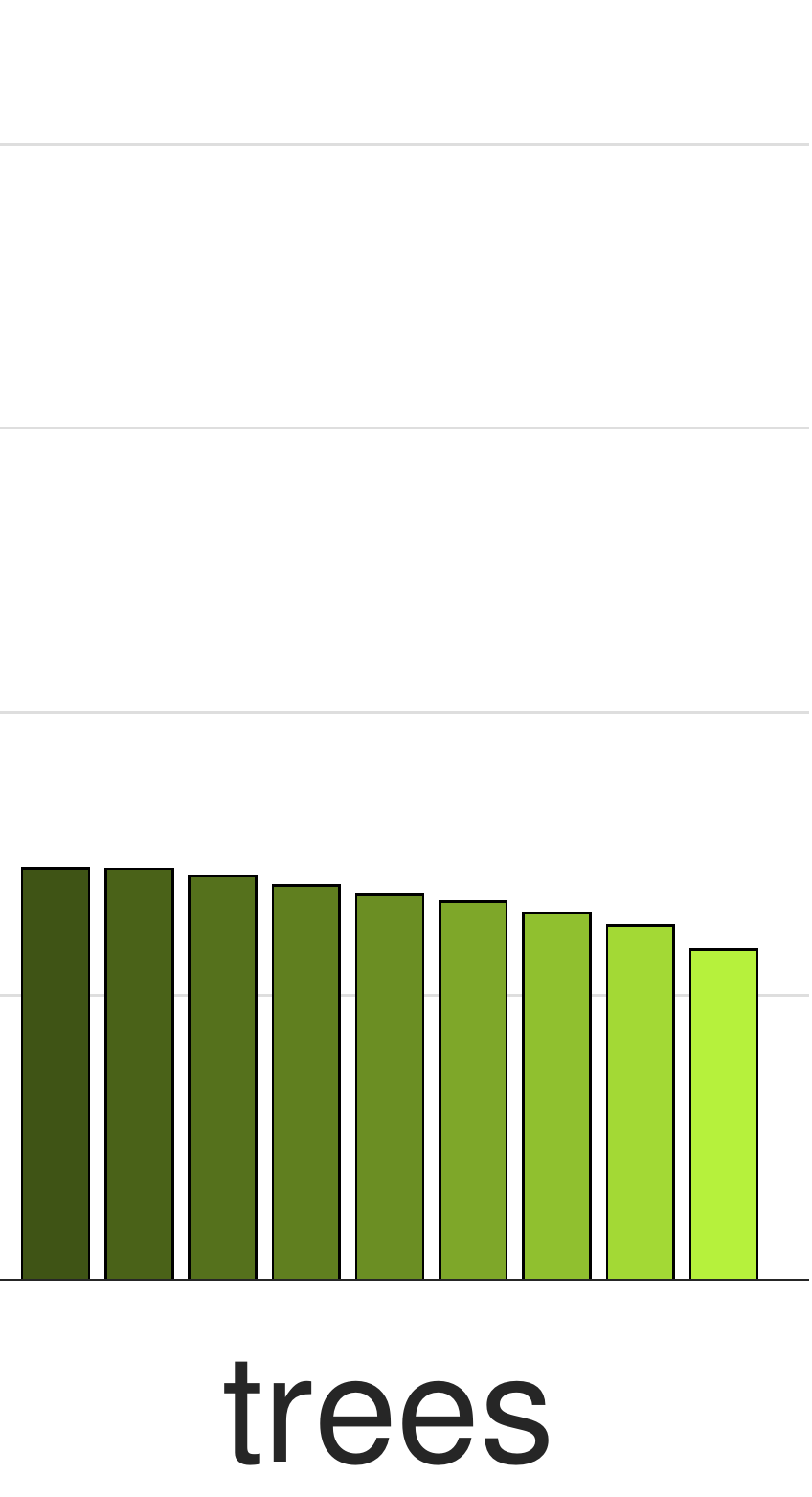} &
      \hspace{-0.45cm}\includegraphics[height=3cm]{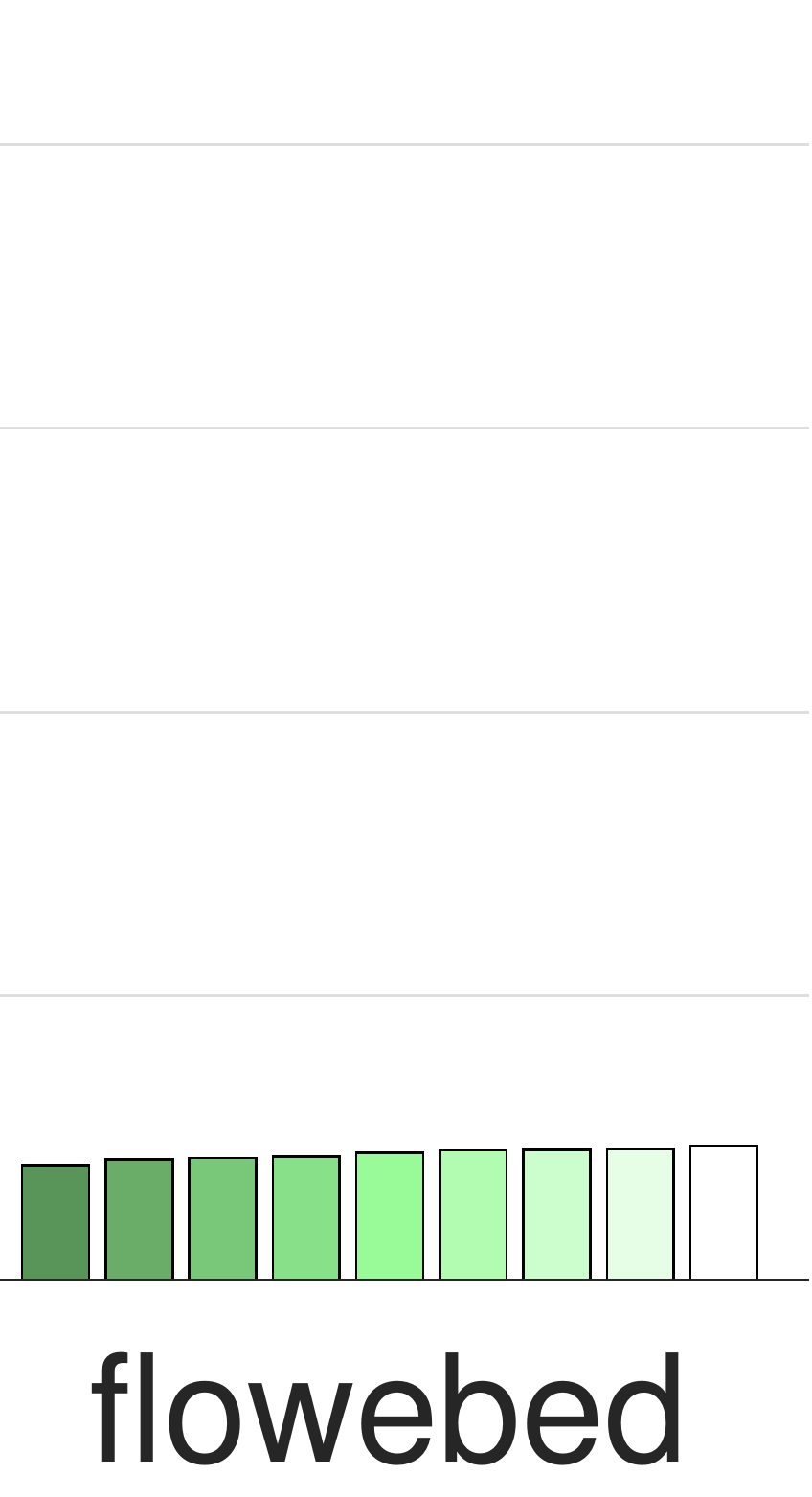} &
      \hspace{-0.45cm}\includegraphics[height=3cm]{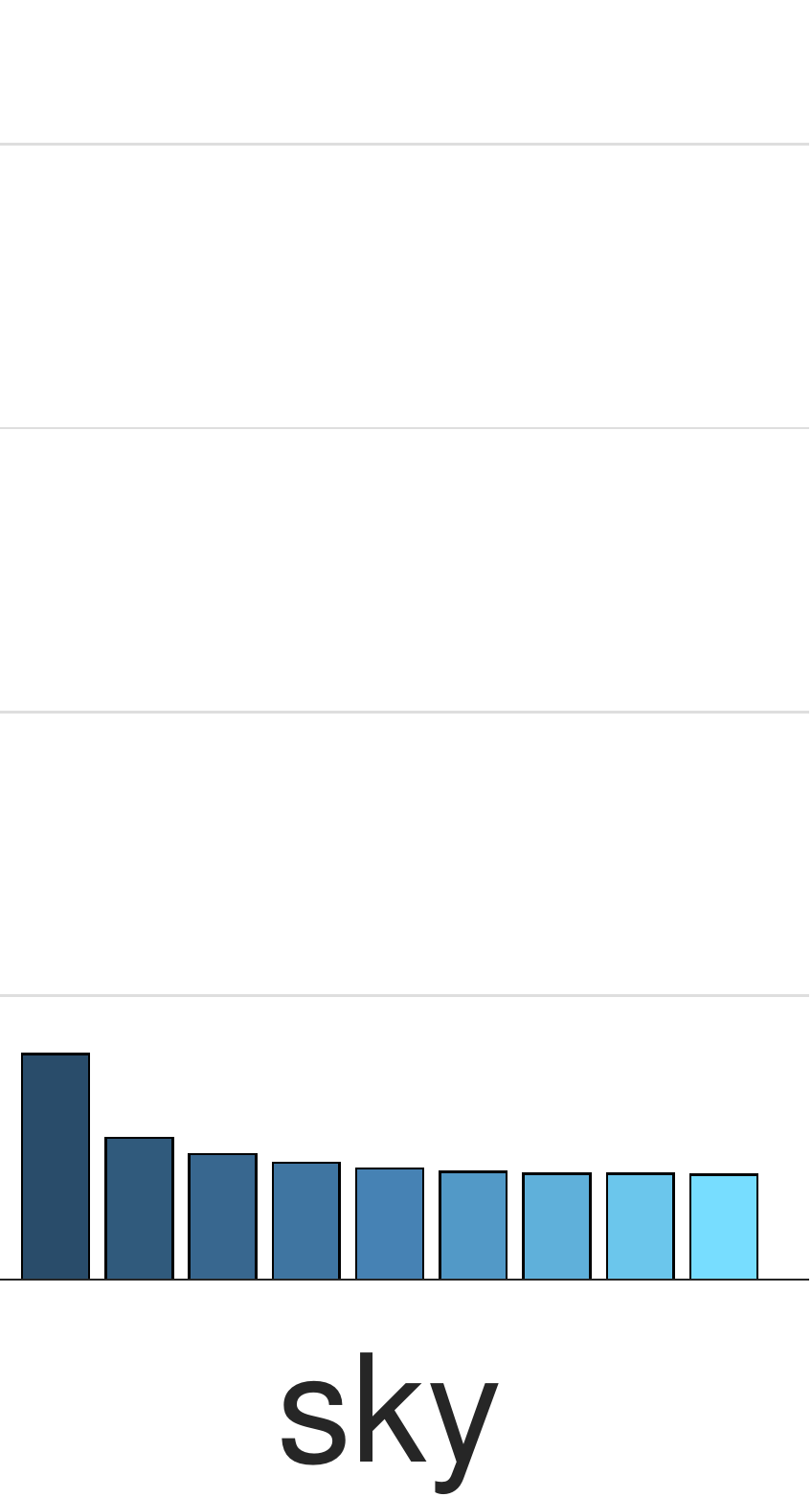} &
      \hspace{-0.45cm}\includegraphics[height=3cm]{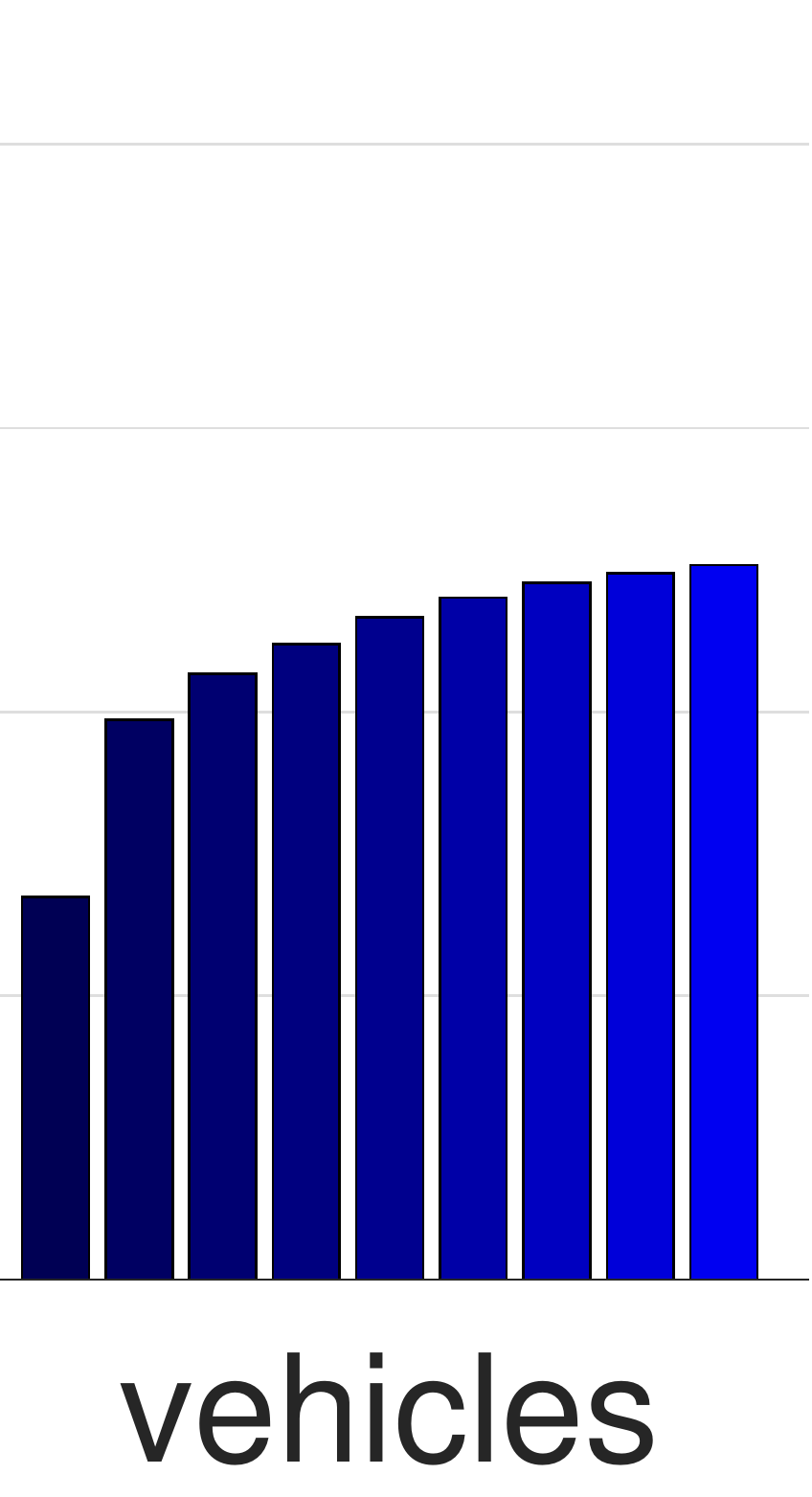} &
                     \includegraphics[height=3cm]{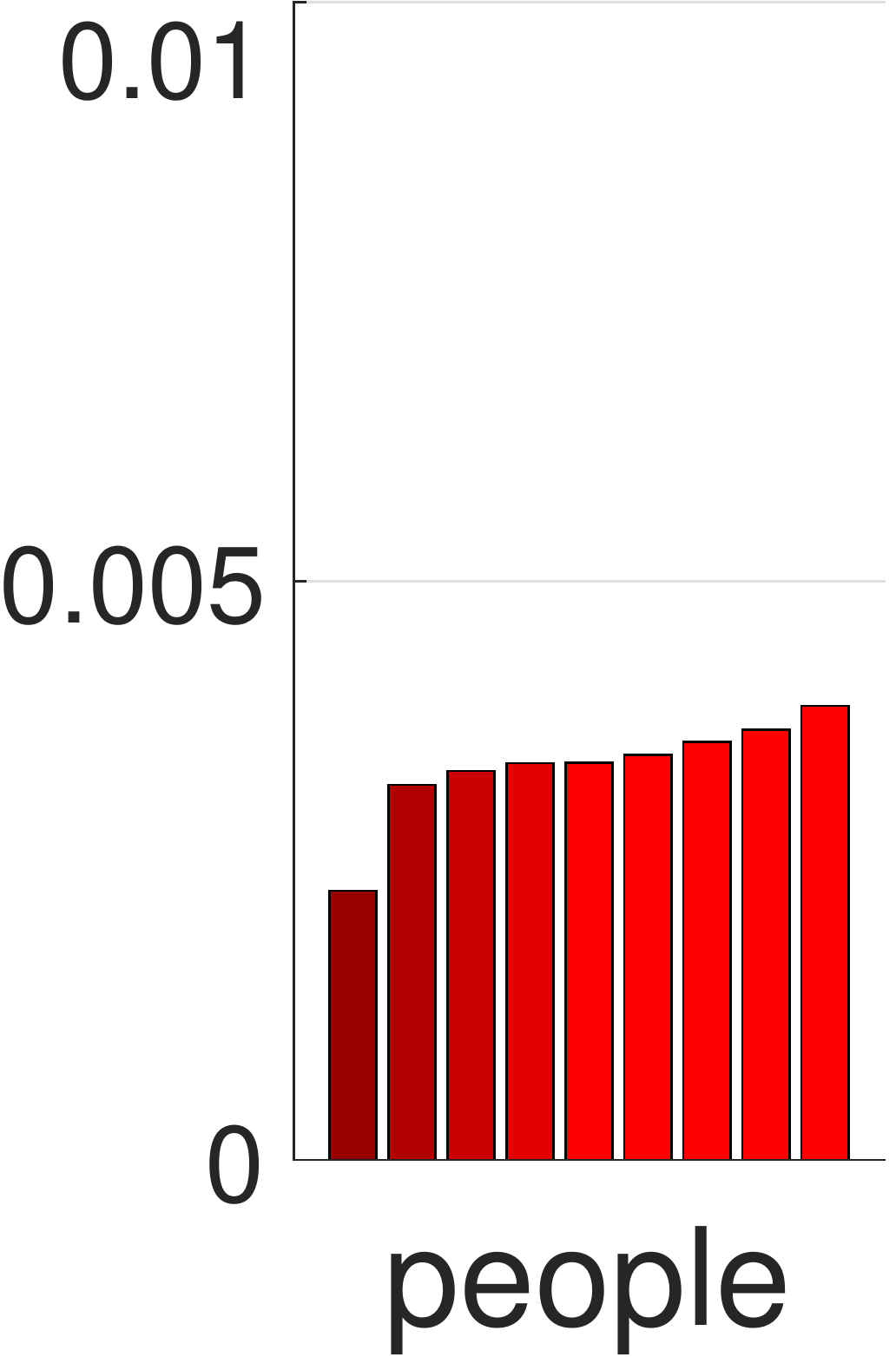} &
      \hspace{-0.45cm}\includegraphics[height=3cm]{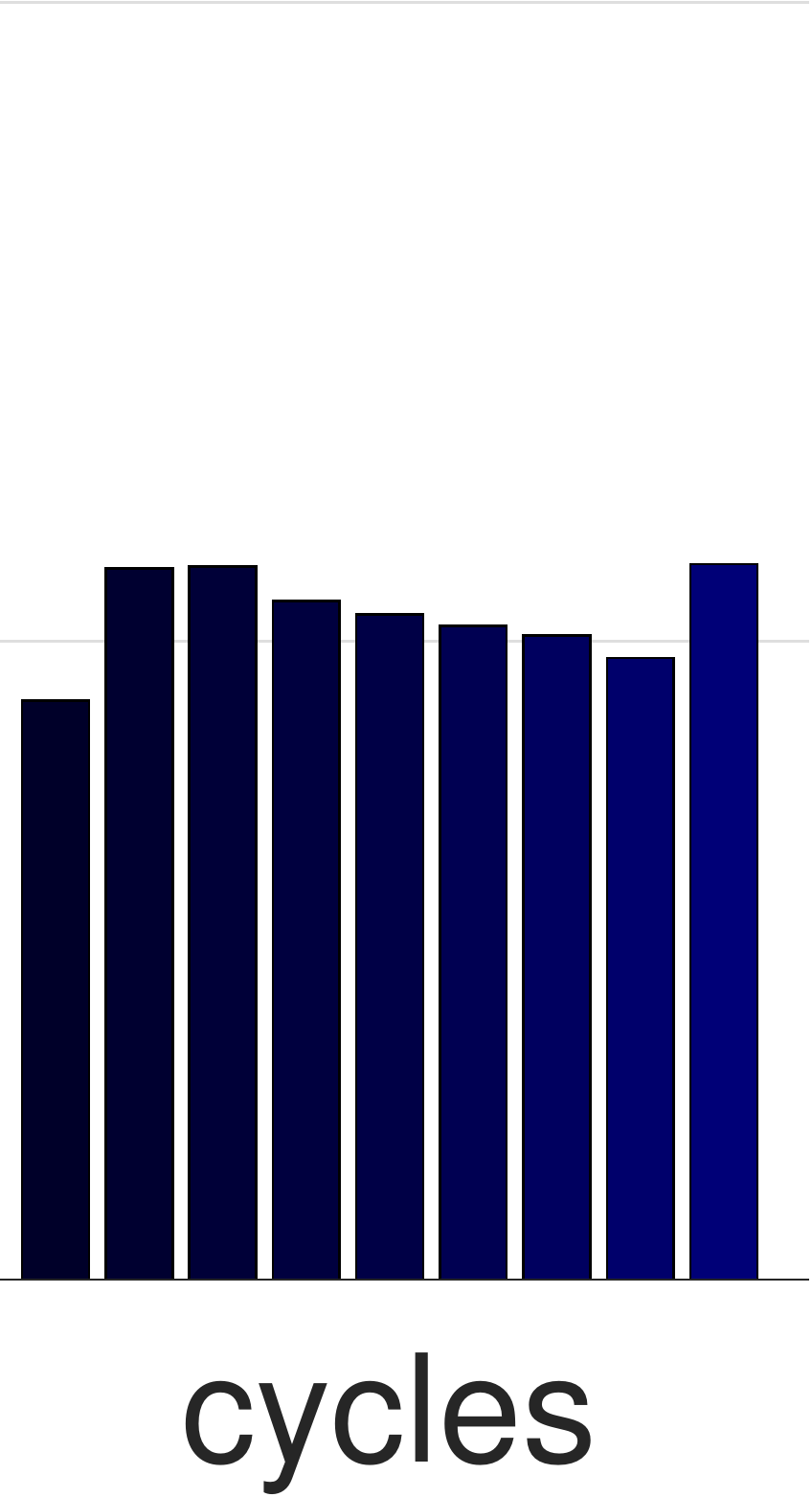}\\
    \end{tabular}
    \caption{Proportion of semantic categories hit by the driver gaze map when thresholded at increasing values (from left to right). Upward trends suggest the element was the focus of the gaze, while downward trends mean the presence of the gaze on the object was only circumstantial. See Sec.~\ref{sec:what} for details.}
    \label{fig:what_gt}
\end{figure*}

The first observation that emerges from the analysis of the dataset is that drivers tend to focus their gaze towards the vanishing point of the road -- often uncaring of close-by road signals, pedestrians walking on a sidewalk or cars coming from the opposite direction. This phenomenon can be qualitatively appreciated in Fig.~\ref{fig:vanishing_point}.
%
This finding can be explained through the notion of \emph{peripheral vision}, introduced in physiological studies to indicate the area around the fixation point where objects and events can still be perceived and interpreted~\cite{sardegna2002encyclopedia}. Accordingly, the largest area of attention is covered by focusing on the vanishing point of the road because, in the task of driving, many of the objects worth of attention have already been perceived when distant.

Additionally, we measured that the average location of the gaze deviates from the road vanishing point as a function of speed and landscape. In fact, due to limited resources our brain compensates the increase of information density - either \emph{spatial} or \emph{temporal} - with a reduction of the visual field size, so that everything that is perceived can also be elaborated~\cite{roge2004influence}.
We have an increase in information \emph{spatial density} in complex scenes, where both many interesting elements and distractors are present. 
Similarly, when the driver travels at higher speed, we have an increase in information \emph{temporal density}, \emph{e.g.} the amount of information we need to elaborate per unit of time.
Contextually to a decreased visual field, the driver needs to move the gaze around the scene more often to capture all the informative elements. Fig.~\ref{fig:gt_across_speed} illustrates how the average gaze position changes at different speeds. 
When the speed increases - from less than 10km/h (a) to more than 70km/h (e) - the average gaze position converges towards the vanishing point of the road, suggesting an increased visual field size able to capture enough of the scene without shifting the gaze. The increased field size is the result of a linear increase of \emph{temporal density} counterbalanced by a stronger decrease of information \emph{spatial density}. In fact, low speed frames occur frequently in downtown sequences where high information \emph{spatial density} is expected, while high speed frames mainly belong to highway driving events, where the amount of information is scarcer. 
The bar plots in Fig.~\ref{fig:gt_across_speed} measure the amount of downtown (red), countryside (green) and highway (blue) frames that concurred to generate the average gaze position for a specific speed range.

\subsection{What do we attend while driving?}
\label{sec:what}
To further investigate the attentional mechanism involved in the driving task, in this section we consider \emph{what} the driver is looking at.
To conduct this analysis we leverage on semantic segmentation - the task of assigning a label to each pixel according to what object it belongs to. We rely on the network proposed by Yu and Koltun~\cite{yu2015multi}, trained on the \texttt{cityscapes}~\cite{Cordts2016Cityscapes} dataset, to recognize the following 10 categories: Road, sidewalk, buildings, traffic lights / signs, trees, road limits, sky, people, vehicles and cycles. Once all frames are labeled we can answer the question of \emph{what is the driver looking at} by measuring the segmentation labels density inside the attentional map\footnote{This analysis comes with the caveat of being dependant on the quality of the semantic segmentation algorithm (which we cannot evaluate on our dataset, if not qualitatively) and the precision of the eye tracking glasses.}, see Fig.~\ref{fig:overview}(d). Since the attentional maps have continuous values in $[0, 1]$, we first build binary maps through linearly spaced thresholds. As the threshold approaches 1, we shrink the observed portion of the scene close to the fixation point. Fig.~\ref{fig:what_gt} shows how different object categories react when such threshold is changed. In particular, an upward trend reveals that a specific category was indeed the focus of the gaze since, as the observed area decreases, the proportion of pixels labeled in that way increases. Interestingly, this is true for traffic lights, road limits, vehicles and people. Similarly, a downward trend reveals that focus on many categories is often only circumstantial, \emph{i.e.} the categories contain objects that happened to be close to what the driver was really looking at. Examples include sidewalks, buildings, trees and sky. 
By fixing a threshold, the plot elucidates on the proportion of observed categories across all the sequences. Our focus is dominated by road and vehicles, but we often observe buildings and trees even if they contain little information useful to drive.

\begin{figure*}[t]
\begin{center}
\includegraphics[width=\textwidth]{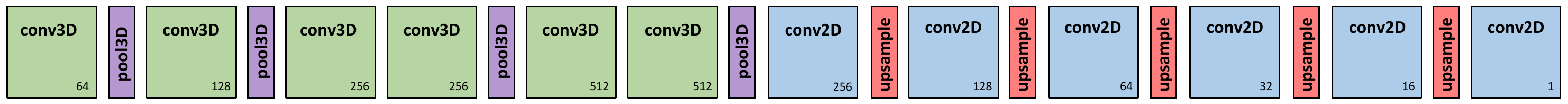}
\end{center}
   \caption{\texttt{COARSE} prediction architecture. The first part of the network performs the feature encoding. The input videoclip is a tensor of size $3\times16\times112\times112$ that undergoes a sequence of conv3D and pool3D layers that gradually squeeze it to size $512\times1\times7\times7$. All conv3D have kernel size (3,3,3) and ReLU activation units; all pool3D have pool size (2,2,2) except the first one that has pool size (1,2,2).  In order to obtain a saliency map with the same spatial size of the input frame, the feature representation is decoded through a series of intertwined layers of conv2D and $\times2$ upsampling on the spatial dimensions. All conv2D have kernel size (3,3) and are followed by leaky ReLU activations with $\alpha =.001$. As a result, the output of the network is a tensor of size $1\times112\times112$, \emph{i.e.} the predicted attentional map.}
\label{fig:encoder_decoder}
\end{figure*}
\begin{figure*}[t]
\begin{center}
\includegraphics[width=1.0\textwidth]{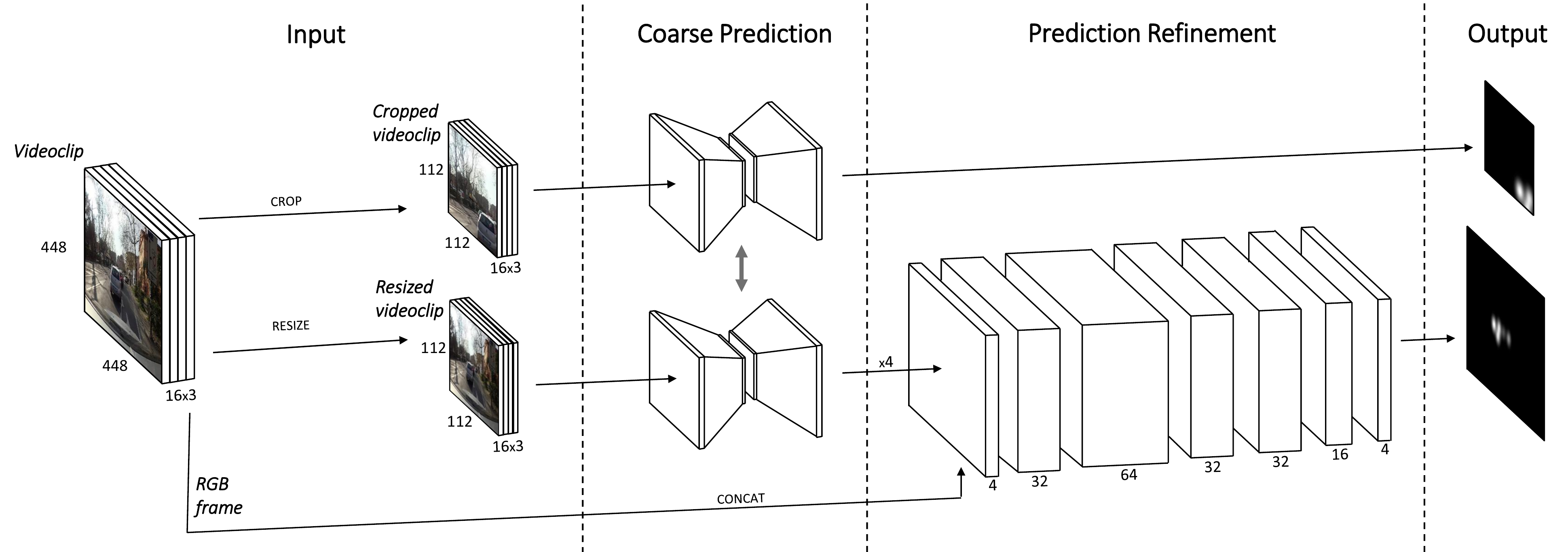}
\end{center}
   \caption{\texttt{COARSE+FINE} prediction architecture. The \texttt{COARSE} module (see Figure \ref{fig:encoder_decoder}) is applied to both a cropped and a resized version of the input tensor, which is a videoclip of 16 consecutive frames. The cropped input is used during training to augment the data and the variety of ground truth attentional maps. The prediction of the resized input is stacked with the last RGB frame of the videoclip and fed to a series of convolutional layers (\texttt{FINE} prediction module) with the aim of refining the prediction. All convolutions have kernel size (3,3). Training is performed end-to-end and weights between \texttt{COARSE} modules are shared. At test time, only the refined predictions are used.}
\label{fig:arch7}
\end{figure*}

\section{Network Models}
\label{sec:models}
In the previous section we studied the driver's focus under a variety of conditions, aiming at providing insights on the mechanisms that govern human attention while driving.
Below, we learn a deep attentional model that, given a driving sequence, is able to focus where the human driver would.

As various recent works show \cite{tran2014learning, karpathy2014large} when dealing with videos, taking explicitly into account the temporal dimension of the input in the network architecture can lead to results that easily outclass the single-frame-input baselines in various high-level video analysis tasks such as video classification and action recognition among others. 
Here, we make the assumption that a short video sequence (\emph{e.g.} half a second) contains enough information to successfully predict where the driver should look in that situation. Indeed, it can be argued that humans take even less time to react to a stimulus. For this reason, we build our approach on 3D convolutions which take as input a fixed-size sequence of 16 consecutive frames from a video (called from now on \emph{videoclip}) and outputs the gaze map for the last frame of the input clip.

\subsection{Coarse gaze prediction module}
The core of our deep network model is a fully convolutional network whose architecture is represented in Fig.~\ref{fig:encoder_decoder}. The first half of the network acts as an encoder and maps the input videoclip in feature space. Conversely, the second block decodes the feature representation in an attentional map which has the same width and height of the input videoclip, but singleton temporal dimension.
In order to perform the encoding, we employ the C3D network by~\cite{tran2014learning} with pre-trained weights and few minor modifications, such as dropping the last convolutional and the fully connected layers in order to maintain spatial information that would be otherwise discarded. See  Fig.~\ref{fig:encoder_decoder} for further details.


During training we resize the training images to $128\times128$ and then randomly crop them to $112\times112$, following the training process described in~\cite{tran2014learning}. Due to the strong bias towards the vanishing point, this standard cropping policy turned out to be insufficient for creating a real variety in the location of the attentional maps; thus the network prediction resulted strongly attracted towards the center of the image.

\subsection{Complete architecture}
In order to solve the aforementioned problem, we moved to an extremely more aggressive crop policy. In this second attempt the training sequences were resized to $256\times256$ before cropping them to $112\times112$. This way the crop constituted less than a quarter of the original image, creating enough variety in the location of the ground truth. This posed however a new challenge: 
As the network was trained on small crops, it learned to predict better localized but significantly wider attentional maps, reflecting the proportion between salient and non-salient areas seen at training time.

We thus enhanced the network architecture towards a more sophisticated multi-input multi-output approach, shown in Fig.~\ref{fig:arch7}. The network still takes a 16 frames videoclip as input, although now there are three different data streams. The first stream provides the model with a randomly cropped videoclip, as explained above. This videoclip passes through the encoding-decoding pipeline and produces a coarse prediction: A first loss on this output is employed to force the model to learn variety in the prediction position and avoid the trivial \emph{hit-the-center} solution.
The second stream feeds the model with the same uncropped videoclip, but resized to match the input shape. This videoclip also goes through the encoding-decoding stack, producing a new coarse uncropped saliency prediction. The prediction is then upsampled and stacked on the RGB image of the last frame of the videoclip, which is provided as third input. Eventually, the concatenated tensor is passed through a last block of convolutions, with the purpose of refining the prediction. This last step also exploits the appearance information of the original videoclip. On the output of this block - our final attentional map prediction - we compute the second loss.\\
%

\noindent In the following experimental sections we evaluate both quantitatively and qualitatively the proposed model. In Sec.~\ref{sec:exp1} we rely on saliency metrics to measure the network's performance against several baselines and state-of-the-art video saliency methods; while in Sec.~\ref{sec:exp2} we analyze how well the model mimics the driver's focus dynamics.

\begin{table*}[t]
\centering
\caption{Training and test results obtained by both the baselines and the proposed networks. See text for details.}
\scalebox{1.173}{
\begin{tabular}{|l||c|c||c|c||c|c|}
\hline
                        & \multicolumn{2}{c||}{Train sequences} & \multicolumn{2}{c||}{Test sequences} & \multicolumn{2}{c|}{CC(GT, MEAN) $<0.3$} \\
                                                & CC               & KL                & CC                & KL                    & CC                & KL             \\ \hline
Baseline (gaussian)    & 0.33 $\pm$ 0.13  &   2.50 $\pm$ 0.63 &   0.33 $\pm$ 0.13 &   2.50 $\pm$ 0.63     & 0.22 $\pm$ 0.15   &   2.70    $\pm$ 0.74      \\ 
Baseline (mean train GT)~~~      & 0.47 $\pm$ 0.28  &   1.65 $\pm$ 1.13 &   0.48 $\pm$ 0.27 &   1.65  $\pm$ 1.17    & 0.17 $\pm$ 0.20   &   2.85    $\pm$ 1.31  \\ \hline
\texttt{COARSE}                    & 0.47 $\pm$ 0.22  &   1.65 $\pm$ 0.96 &   0.44 $\pm$ 0.23 &   1.73  $\pm$ 1.00    & 0.19 $\pm$ 0.18   &   2.74    $\pm$ 1.07  \\ 
\texttt{COARSE+FINE}               & 0.62 $\pm$ 0.26  &   1.21 $\pm$ 0.99 &   0.55 $\pm$ 0.28 &   1.42  $\pm$ 1.07    & 0.30 $\pm$ 0.24   &   2.24    $\pm$ 1.10  \\ \hline
\end{tabular}
}
\label{tab:pred_results}
\end{table*}


\vspace{0.3cm}
\section{Experiments}
\label{sec:exp1}
In this section we quantitatively measure the performance of the proposed \texttt{COARSE} and \texttt{COARSE+FINE} models against different baselines. Following the guidelines in~\cite{bylinskii2016different}, for the evaluation phase we rely on Pearson's Correlation Coefficient (CC) and Kullback--Leibler divergence (KL) measures.\\

\noindent {\bf Training details.} The encoding half of the \texttt{COARSE} network is initialized with pre-trained weights~\cite{tran2014learning}. Training sequences are randomly mirrored to augment the data. End-to-end training minimizes the Mean Squared Error for both losses of the \texttt{COARSE+FINE} model; we employ Adam optimizer with parameters suggested in the original paper~\cite{kingma2014adam}. Train and test are split according to~\cite{alletto2016dr} and 500 central frames from each training sequence
are used for validation.

\subsection{Keeping it simple: Baselines from saliency}

\label{sec:baselines}
\begin{figure}[t]
    \centering
    \begin{tabular}{cc}
        \hspace{-0.2cm}\includegraphics[width=0.24\textwidth]{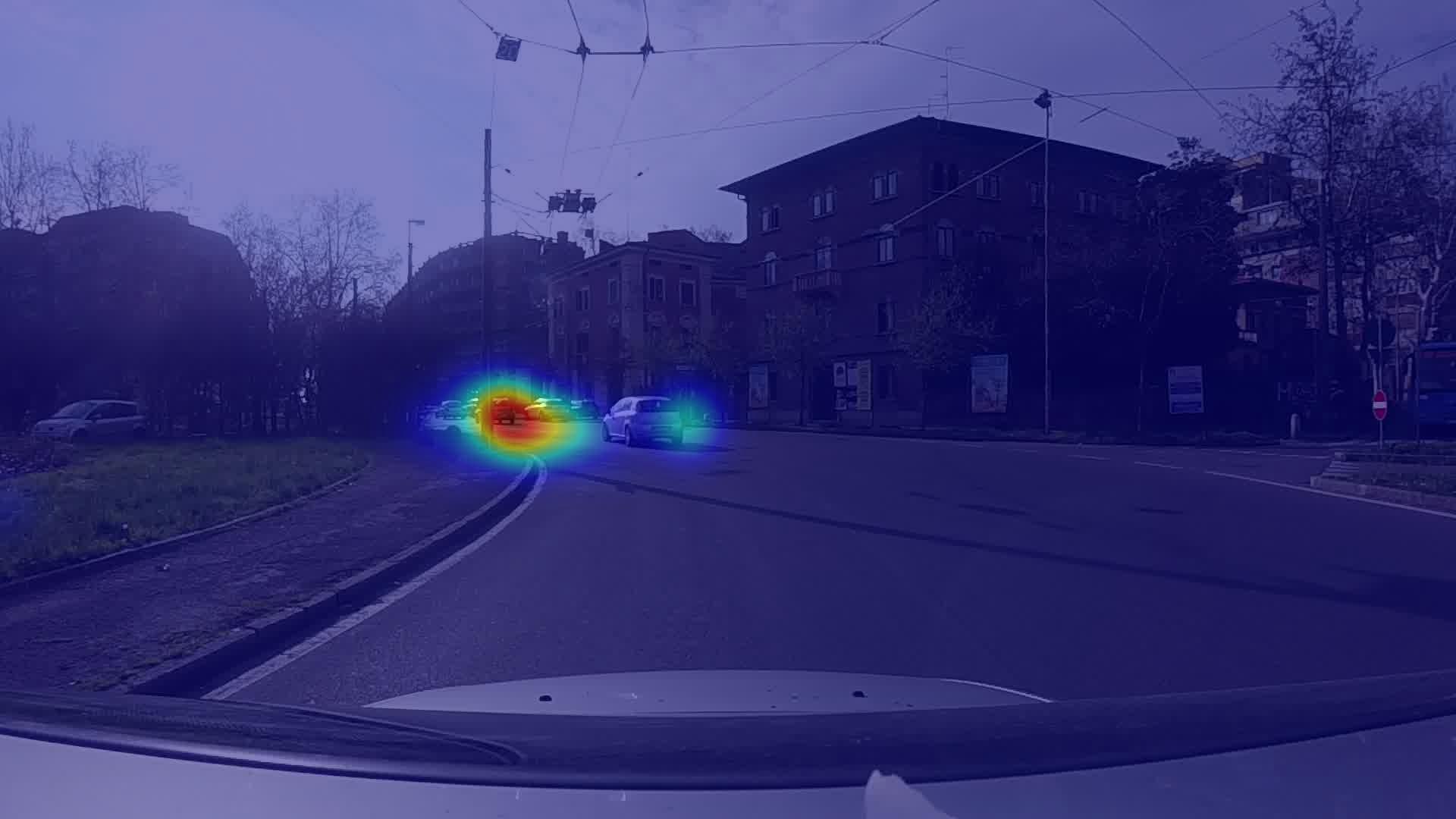} &\hspace{-0.3cm}\includegraphics[width=0.24\textwidth]{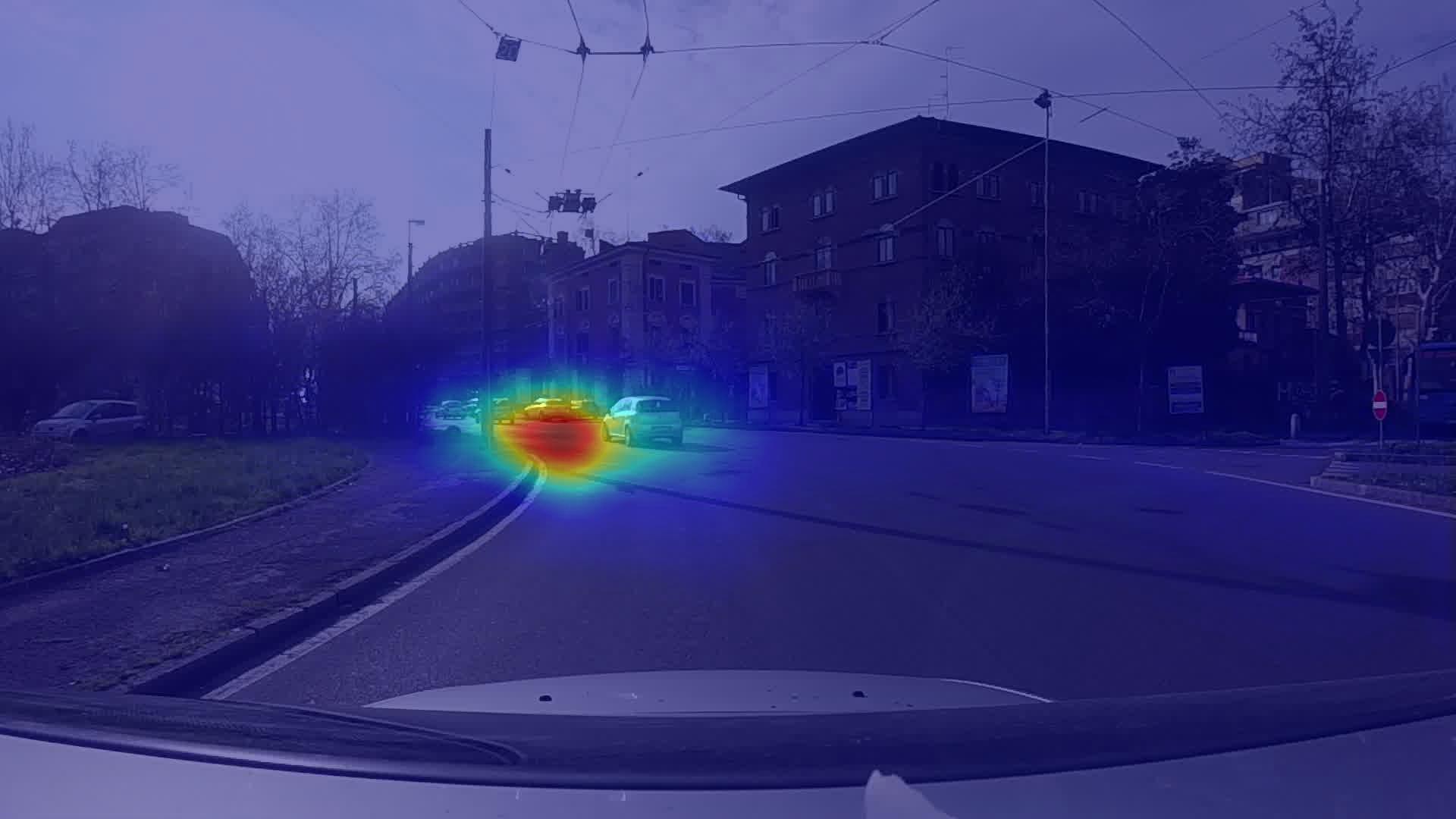}\\
        \hspace{-0.2cm}(a) ground truth & \hspace{-0.3cm}(b) network prediction \\
        \hspace{-0.2cm}\includegraphics[width=0.24\textwidth]{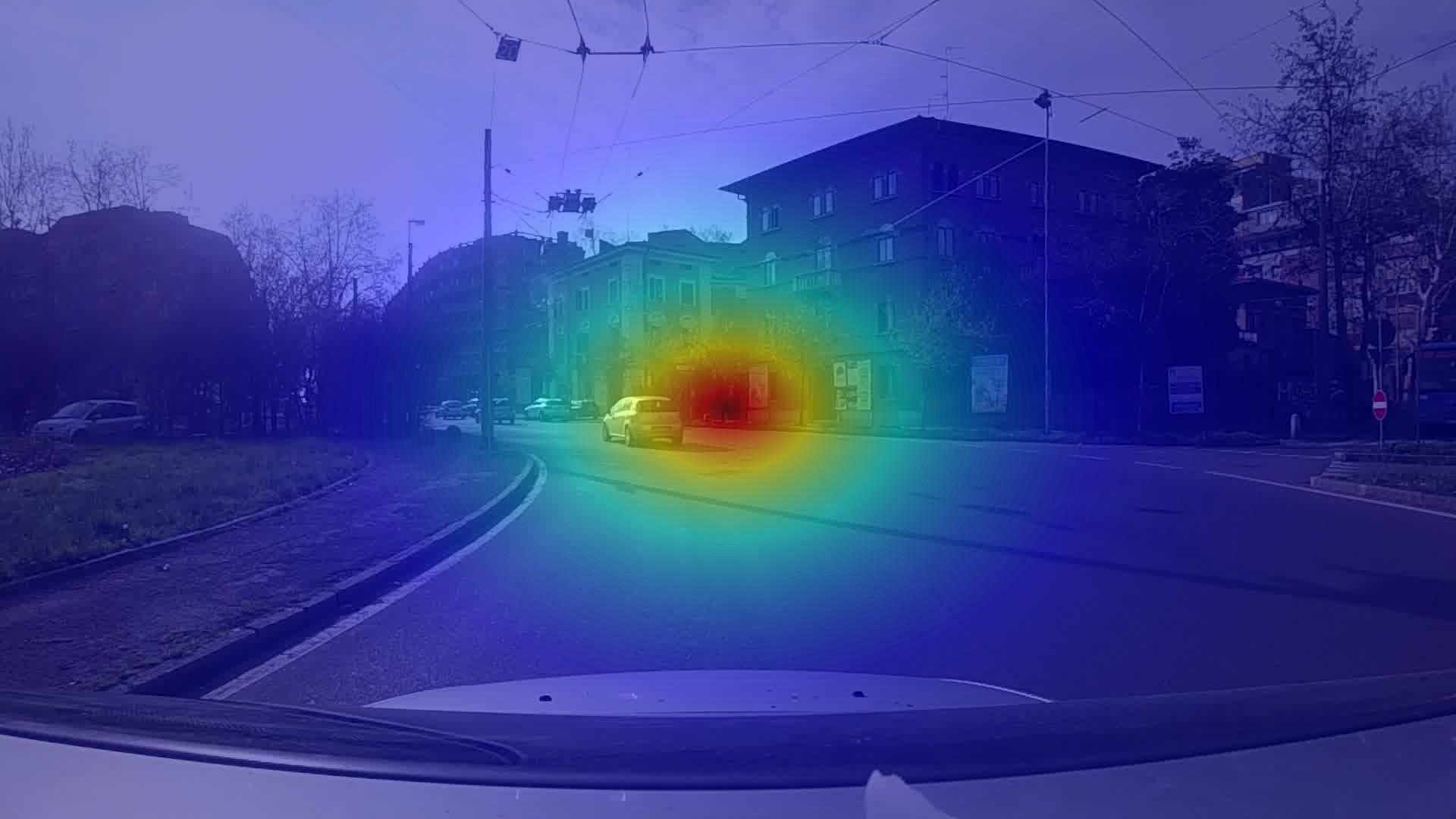} &\hspace{-0.3cm}\includegraphics[width=0.24\textwidth]{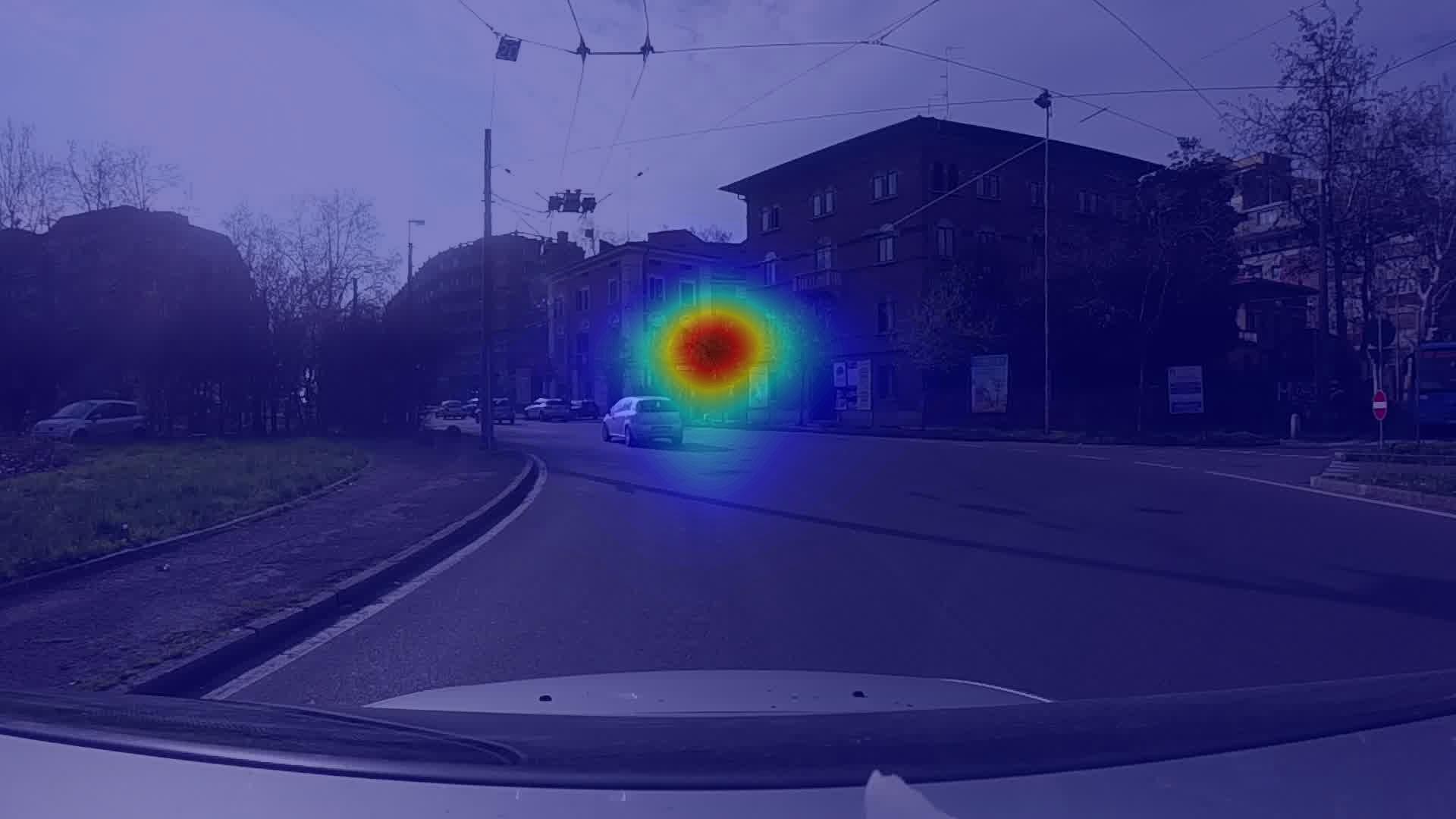}\\
        \hspace{-0.2cm}(c) baseline: gaussian & \hspace{-0.3cm}(d) baseline: average GT
    \end{tabular}
    \caption{Comparison between the ground truth, the prediction and the baselines for an example frame.}
    \label{fig:baseline_vs_prediction}
\end{figure}

It is widely known that a centered Gaussian, stretched to the aspect ratio of the image, makes for an incredibly effective baseline for the visual saliency task. This static baseline scores better than many methods benchmarked on the \texttt{MIT300}~\cite{mit-saliency-benchmark} dataset. Section~\ref{sec:where} revealed that a similar bias affects the \drive dataset (see Fig.~\ref{fig:vanishing_point}). Thus, to validate the proposed model, we compare it against both the aforementioned baseline and a more task-driven version of it built as the average of all training set attentional maps, Fig.~\ref{fig:baseline_vs_prediction}(c) and (d). 
Results are reported in Tab.~\ref{tab:pred_results}: The second baseline performs better than both the Gaussian baseline and the \texttt{COARSE} model on the test set, but significantly worse than \texttt{COARSE+FINE} model. Indeed the latter has a relative gain of about $+25\%$ over the \texttt{COARSE} model and about $+15\%$ over the average training ground truth.

\subsection{Comparison with state-of-the-art}
\label{sec:comparison}
In Tab.~\ref{tab:comparison} we report results from two recent unsupervised video saliency methods~\cite{wang2015saliency, wang2015consistent} and a supervised one~\cite{6942210} on the test set. Both unsupervised methods rely on appearance and motion discontinuities and are easily fooled by the motion of the roof-mounted camera. Unfortunately,~\cite{6942210} is trained on the \emph{Action in The Eye} dataset and thus performs poorly on this domain. As far as we know, no supervised video saliency method releases the source code allowing us to re-train it, nor reports performance on \texttt{DR(eye)VE.}
Results shown in Tab.~\ref{tab:comparison} call for supervised methods aware of both the semantic of the scene and the peculiarities of the task. To our knowledge, our proposal is the first deep model for driving attention, and video saliency to a greater extent.

\begin{table}[t]
    \centering
    \caption{Comparison with video saliency state-of-the-art methods.}
    \scalebox{1.21}{
        \begin{tabular}{|l|c|c|}
        \hline
                & CC & KL \\ \hline
        Wang \emph{et al.} \cite{wang2015consistent} \quad\quad\quad\quad& 0.08 $\pm$ 0.11  & 3.77 $\pm$ 0.77    \\ \hline
        Wang \emph{et al.} \cite{wang2015saliency}    & 0.03 $\pm$ 0.09  &  4.24 $\pm$ 1.13   \\ \hline
        Mathe \emph{et al.} \cite{6942210}    & 0.04 $\pm$ 0.08  &  3.92 $\pm$ 0.53   \\ \hline
        \end{tabular}
    }
    \label{tab:comparison}
\end{table}

\subsection{New annotations to escape the bias}
\label{sec:new_annotations}
Despite showing good results, the baselines introduced in Sec.~\ref{sec:baselines} are of no interest for the driving task as they are not able to generalize when required.
There is a strong unbalance between lots of trivial-to-predict scenarios of little interest and few but important hard-to-predict events.
To enable the evaluation of our model under such circumstances, we select from the \drive dataset those sub-sequences whose ground truth poorly correlates with the average ground truth of the whole sequence ($\text{CC}<0.3$), under the assumption that in these situations something worth noticing is drifting the driver's focus from the vanishing point of the road.
The last column of Tab.~\ref{tab:pred_results} reports the results computed on such subset. When tested on these sequences, the Gaussian baseline outperforms both the \texttt{COARSE} model and the average training ground truth baseline. To interpret this result, consider that when measuring a distance between distributions, a high probability but wrong prediction is severely penalized over a somehow uncertain prediction (\emph{i.e.} a Gaussian with high variance). Nonetheless, the \texttt{COARSE+FINE} model scores higher than all other methods with a relative gain of about $+35\%$ over the second best.

\begin{figure*}[ht!]
    \centering
    \begin{tabular}{ccccc}
    \includegraphics[width=0.18\textwidth]{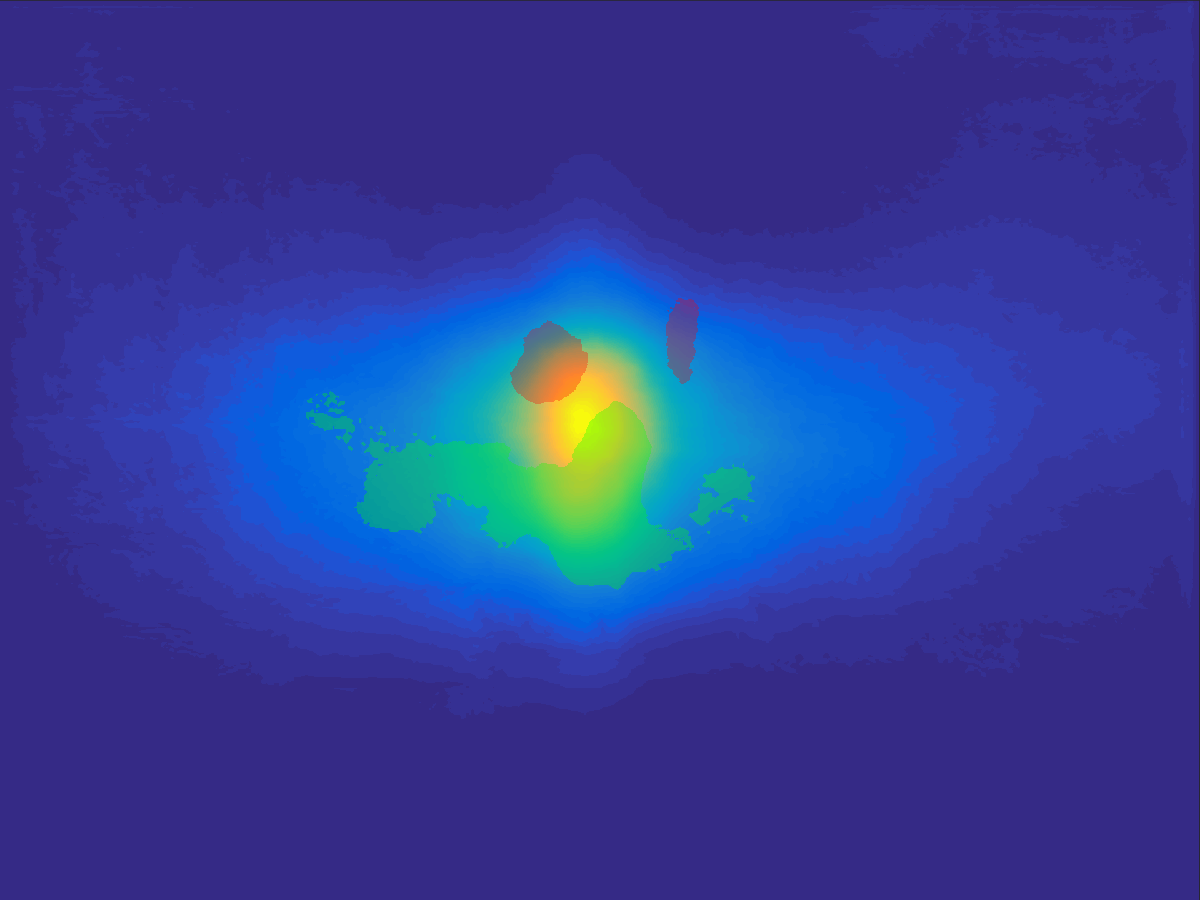} &
    \hspace{-0.2cm} \includegraphics[width=0.18\textwidth]{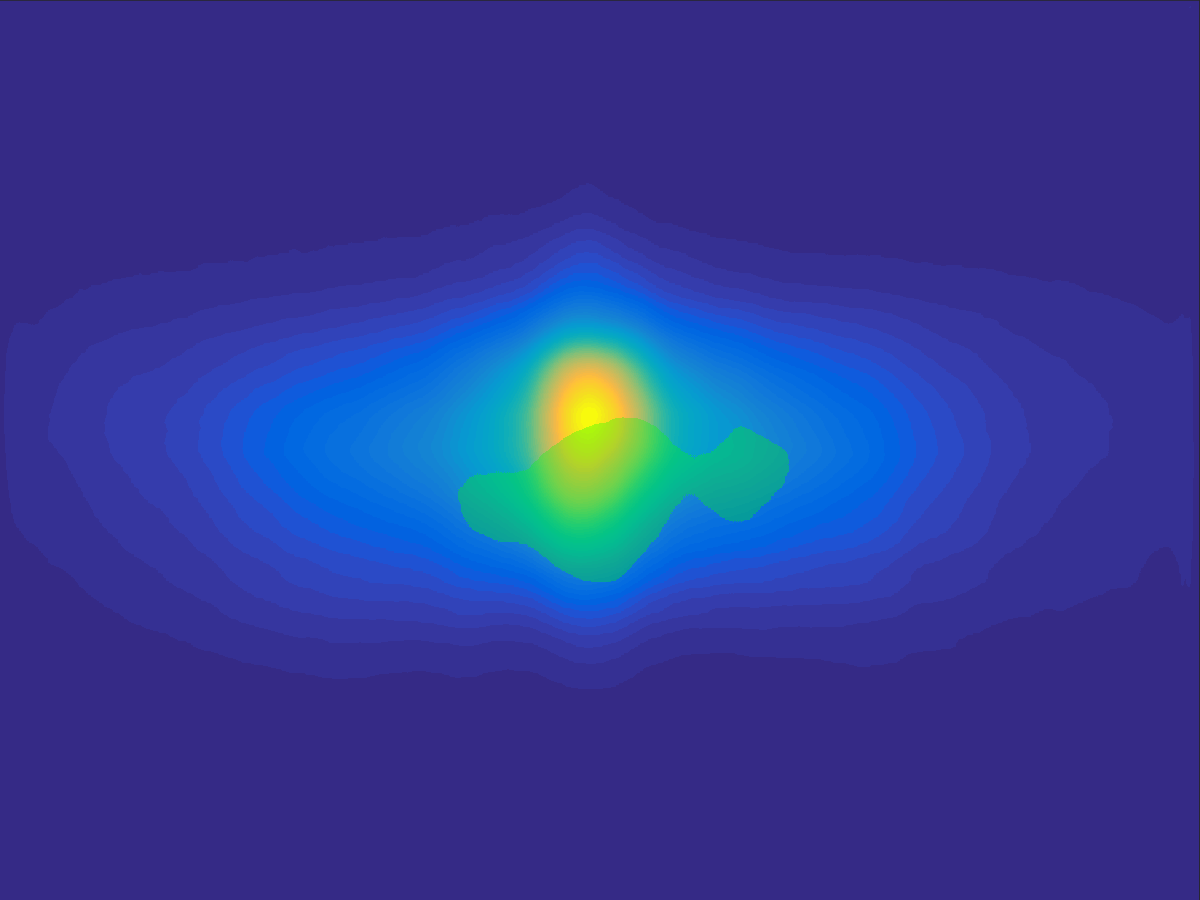} &
    \hspace{-0.2cm} \includegraphics[width=0.18\textwidth]{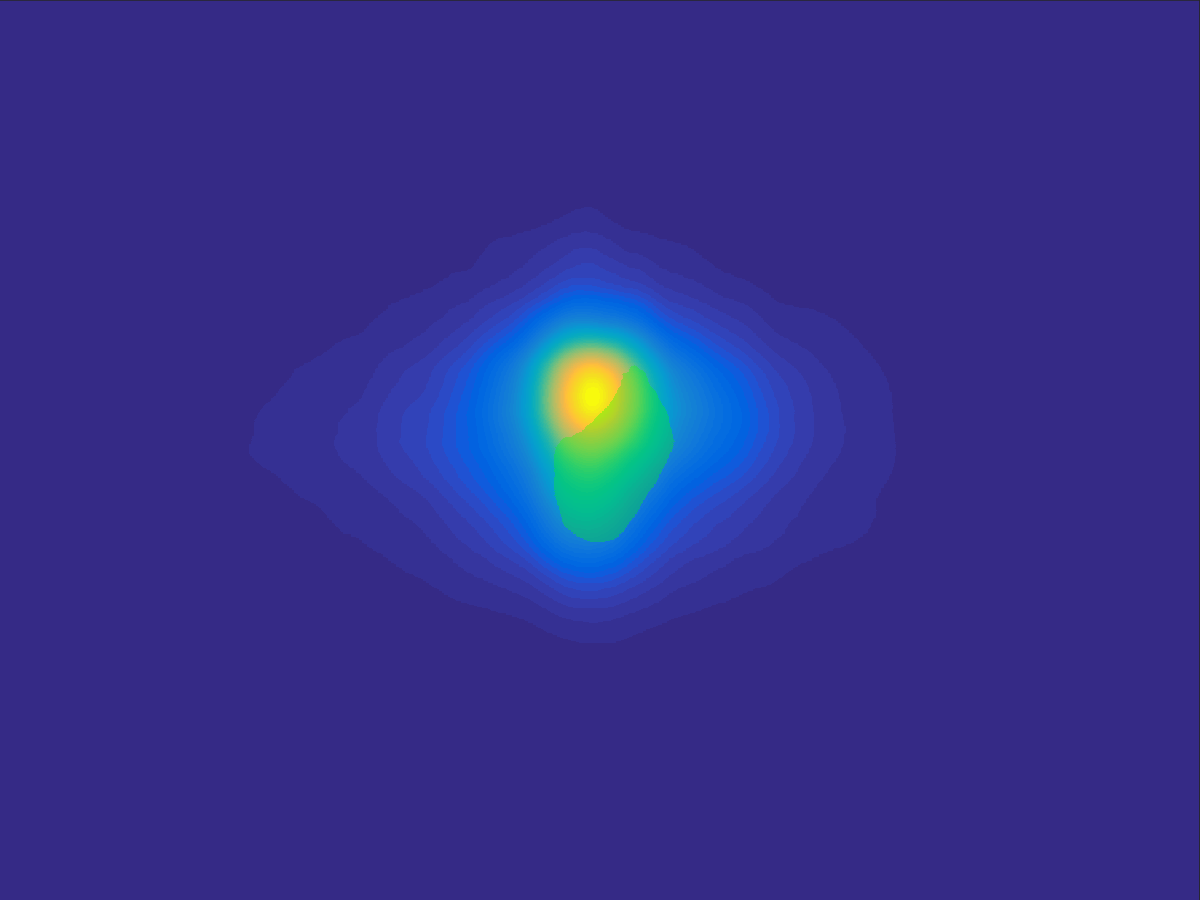} &
    \hspace{-0.2cm} \includegraphics[width=0.18\textwidth]{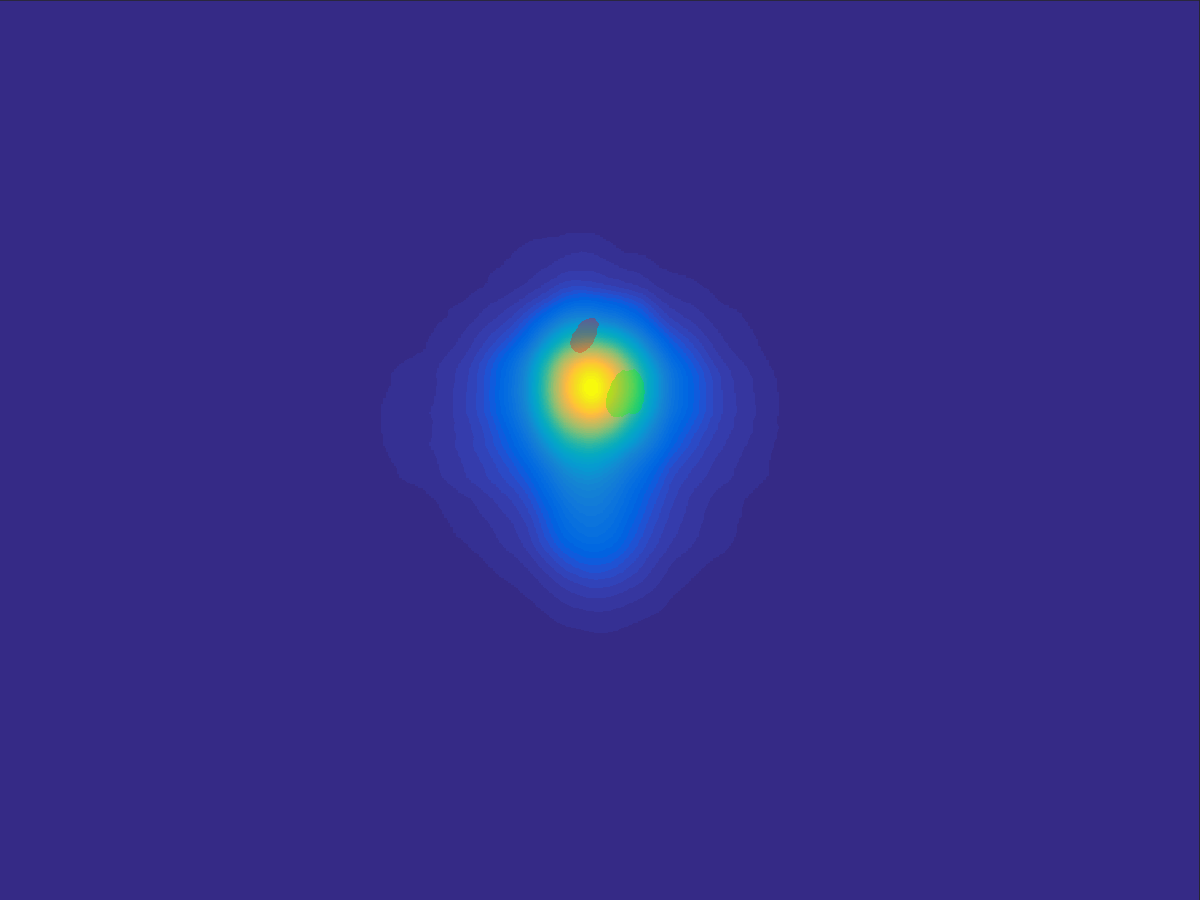} &
    \hspace{-0.2cm} \includegraphics[width=0.18\textwidth]{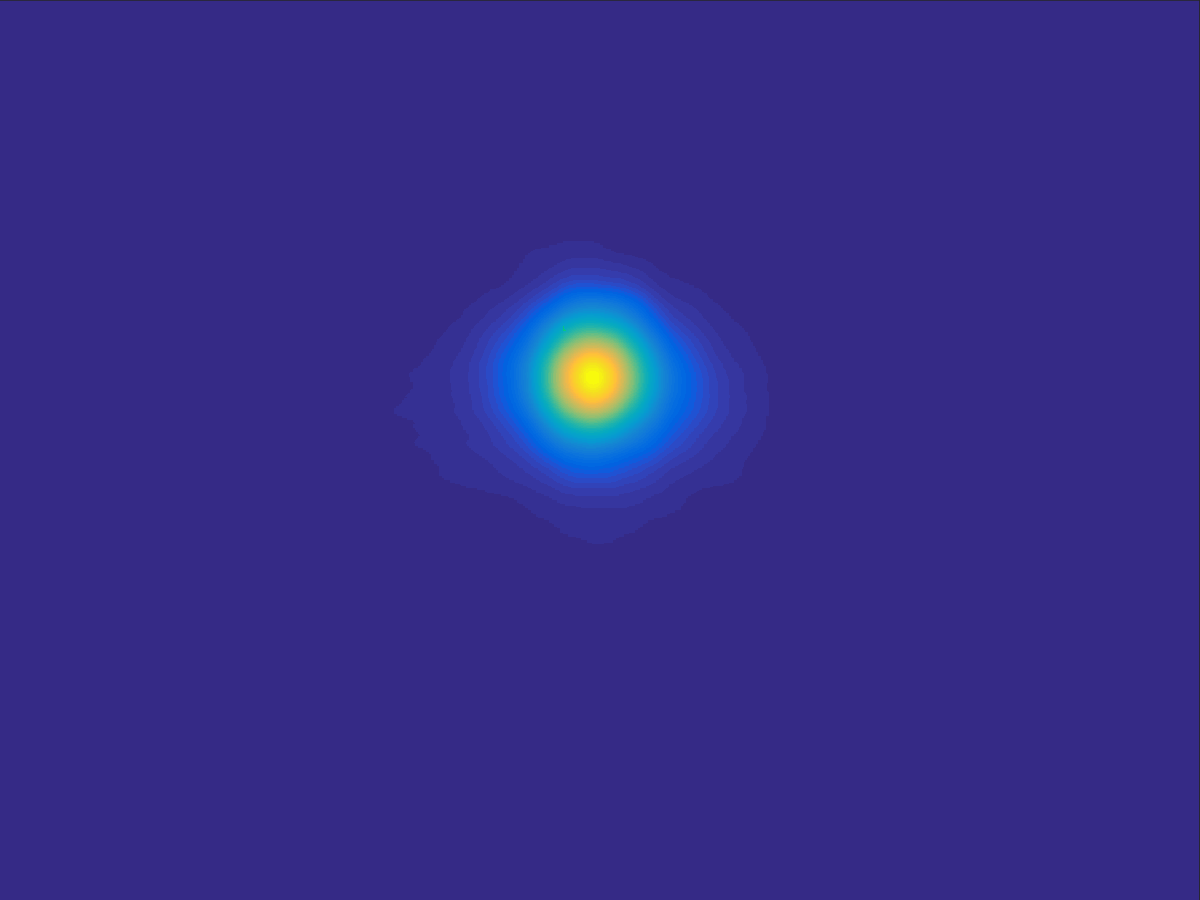}\\
    (a) $0\leq \mathrm{km/h}\leq10$ &
    (b) $10\leq \mathrm{km/h}\leq30$ &
    (c) $30\leq \mathrm{km/h}\leq50$ &
    (d) $50\leq \mathrm{km/h}\leq70$ &
    (e) $70\leq \mathrm{km/h}$
    \end{tabular}
    \hspace{0.2cm}
    \caption{Model prediction averaged across all test sequences and grouped by driving speed. We highlight areas in which the mean prediction deviates by more than 10\% from the mean ground truth. Precision errors are overlaid in green, while recall error in red.}
    \label{fig:pred_across_speed}
\end{figure*}


\section{Do we capture the attentional dynamics?}
\label{sec:exp2}

In Sec.~\ref{sec:exp1} we quantitatively evaluated the proposed network. Here, we qualitatively investigate the ability of the model to learn both \emph{where} and \emph{what} a human driver would focus while driving. The results are then compared against the analysis previously introduced in Sec.~\ref{sec:dreyeve_dataset_analysis}.

Figure~\ref{fig:pred_across_speed} shows the average attentional maps predicted by our model, arranged by speed range. On each plot we also overlay precision errors (green) and recall errors (red), \emph{i.e.} pixels whose value differs by more than $10\%$ from the analogous ground truth plots reported in Fig~\ref{fig:gt_across_speed}. We observe that i) the model generally succeeds in capturing the location of the driver gaze at different speed, ii) errors are mostly due to precision (prediction is wider than ground truth) and iii) errors decrease as the speed increases, as the lower variance of the gaze at high speed makes the modeling task easier.

\begin{figure}[ht!]
    \centering
    \includegraphics[width=1.0\columnwidth]{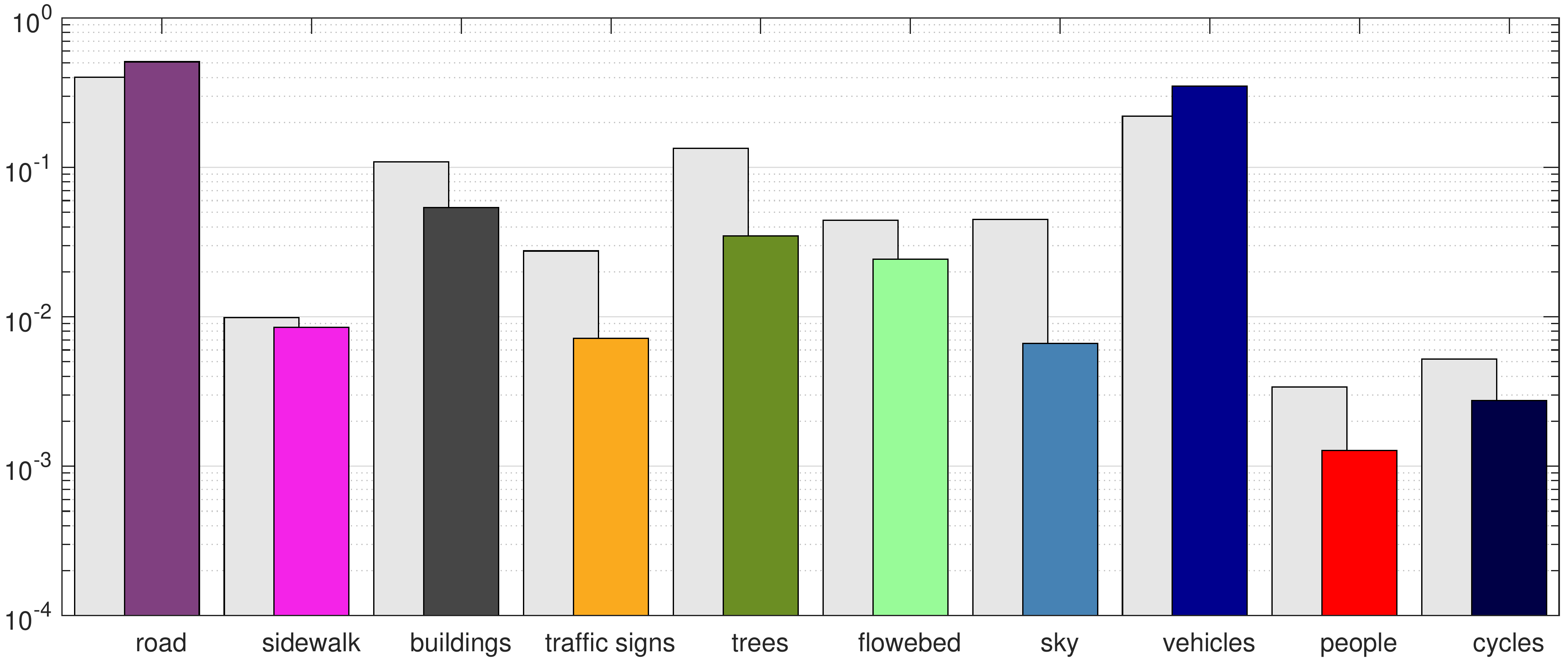}
    \caption{Comparison between ground truth (gray bars) and predicted attentional maps (colored bars) when used to mask semantic segmentation of the scene. Absolute errors exist, but the two bar series agree on the relative importance of different categories.}
    \label{fig:semantic_prediction}
\end{figure}

We repeat the analysis of Sec.~\ref{sec:what} employing our predictions to determine how often our model focuses on different object categories. 
Such comparison is shown in Fig.~\ref{fig:semantic_prediction} where values are reported on a log scale in order to highlight differences for low populated categories. Gray bars represent the ground truth reference values averaged across different thresholds (see Sec.~\ref{sec:what} for details), while prediction results bars are colored based on the semantic segmentation color map introduced in Fig.~\ref{fig:overview}(d). According to Fig.~\ref{fig:semantic_prediction}, some level of absolute error is found for all categories. Yet, our model was able to coarsely recognize the relative importance of different categories, \emph{e.g.} road and vehicles still dominate the list, while people and cycles are at the bottom. This correlation is confirmed by Kendall Rank Coefficient, which scored $51.11\%$ when computed on the two bar series.

\section{Conclusions}
In this work we investigated the spatial and semantic attentional dynamics of the human driver and designed a deep network able to replicate such behavior while driving.
These results eventually pave the way for a new assisted driving module, where real-time attentional maps support the driver by both decreasing fatigue and helping in keeping focus. We argue that such attentional maps can be less invasive than other Advanced Driver Assistance Systems that directly act on the car (\emph{e.g.} by activating breaks), whereas attentional maps leave full control to the driver.


\bibliographystyle{IEEEtran}
\bibliography{IEEEabrv,bibliography}

\end{document}